\documentclass[10pt,twocolumn,letterpaper]{article}

\usepackage{wacv}
\usepackage{times}
\usepackage{epsfig}
\usepackage{graphicx}
\usepackage{amsmath}
\usepackage{amssymb}
\usepackage{float}
\usepackage{tabularx, enumitem,booktabs, cellspace, multirow}
\usepackage{subcaption}
\usepackage{footnote}

\wacvfinalcopy % *** Uncomment this line for the final submission

\def\wacvPaperID{1071} % *** Enter the wacv Paper ID here

\ifwacvfinal\fi
\begin{document}

%%%%%%%%% TITLE
\title{Global Table Extractor (GTE): A Framework for Joint Table Identification and Cell Structure Recognition Using Visual Context}
%\begin{comment}
% \titlerunning{ECCV-20 submission ID \ECCVSubNumber} 
% \authorrunning{ECCV-20 submission ID \ECCVSubNumber} 
% \author{Anonymous ECCV submission}
% \institute{Paper ID \ECCVSubNumber}
%\end{comment}
%******************

% \begin{comment}

\author{Xinyi Zheng\\
University of Michigan\\
% TODO Add footnote
{\tt\small zxycarol@umich.edu}
\and
Douglas Burdick\hspace{2cm} Lucian Popa\\
IBM Research-Almaden\\
{\tt\small \{drburdic, lpopa\}@us.ibm.com}
\and
Xu Zhong\\
IBM Research Australia\\
{\tt\small peter.zhong@au1.ibm.com}\and
Nancy Xin Ru Wang\\
IBM Research-Almaden\\
{\tt\small wangnxr@ibm.com}
% Lucian Popa\inst{1}\orcidID{0000-0002-0659-9144} \and 
% \\
% Peter Zhong \and 
% Nancy Xin Ru Wang\inst{1}\orcidID{0000-0002-3041-6934} 
}

% Authors at the same institution
%\author{First Author \hspace{2cm} Second Author \\
% Authors at different institutions
% \author{First Author \\
% \and
% Second Author \\

\maketitle
\ifwacvfinal\thispagestyle{empty}\fi

%%%%%%%%% ABSTRACT
\begin{abstract}
% \vspace{-0.1cm}
Documents are often used for knowledge sharing and preservation in business and science, within which are tables that capture most of the critical data. 
Unfortunately, most documents are stored and distributed as PDF or scanned images, which fail to preserve logical table structure. 
Recent vision-based deep learning approaches have been proposed to address this gap, but most still cannot achieve state-of-the-art results.
We present \emph{Global Table Extractor (GTE)}, a vision-guided systematic framework for joint table detection and cell structured recognition, which could be built on top of any object detection model. With \emph{GTE-Table}, we invent a new penalty based on the natural cell containment constraint of tables to train our table network aided by cell location predictions. 
\emph{GTE-Cell} is a new hierarchical cell detection network that leverages table styles.
Further, we design a method to automatically label table and cell structure in existing documents to cheaply create a large corpus of training and test data. We use this to enhance PubTabNet with cell labels and create FinTabNet, real-world and complex scientific and financial datasets with detailed table structure annotations to help train and test structure recognition.
Our framework surpasses previous state-of-the-art results on the ICDAR 2013 and ICDAR 2019 table competition in both table detection and cell structure recognition. %with a significant 5.8\% improvement in the full table extraction system. 
Further experiments demonstrate a greater than 45\% improvement in cell structure recognition when compared to a vanilla RetinaNet object detection model in our new out-of-domain FinTabNet. 
\end{abstract}

% %%%%%%%%% BODY TEXT
\section{Introduction}

\begin{figure}[t]
\centering
   \includegraphics[width=0.4\textwidth]{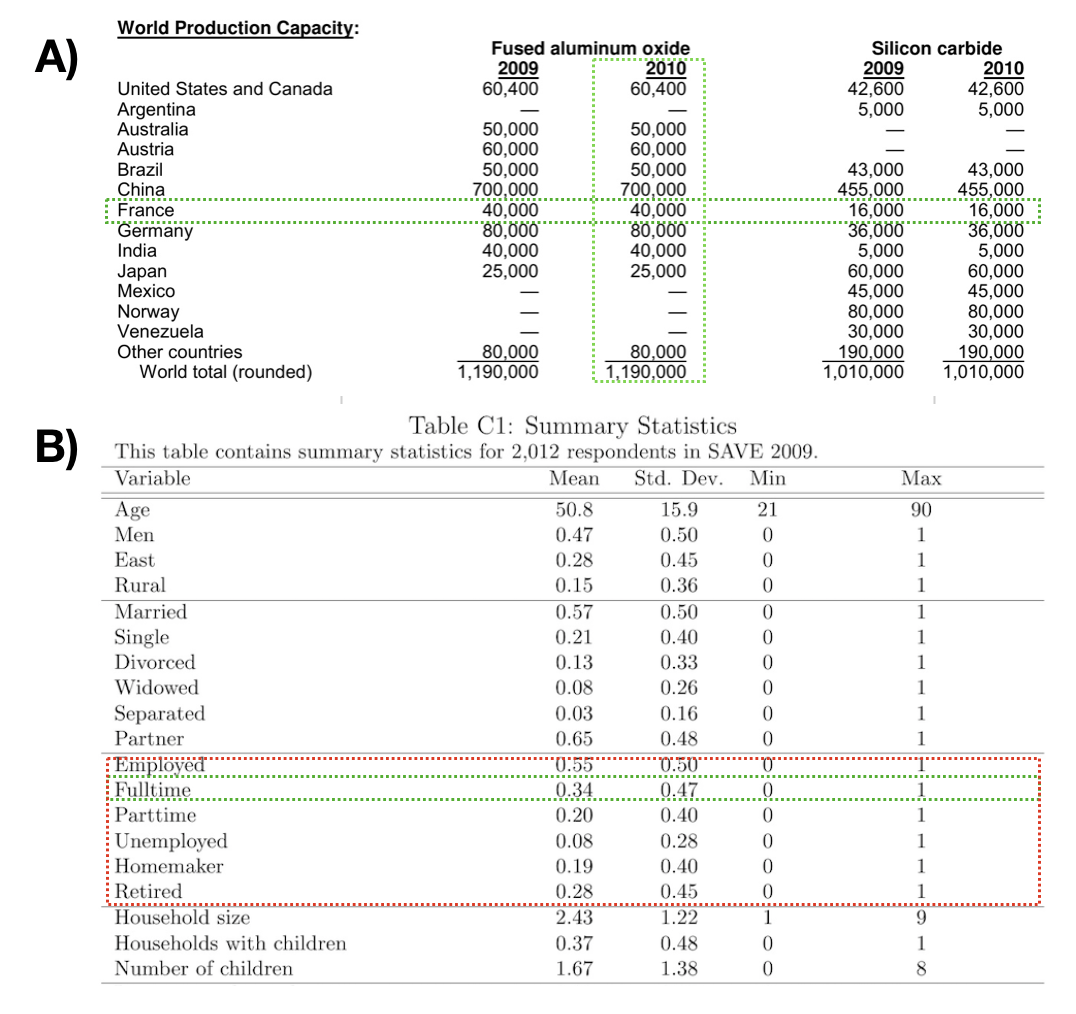}
    \caption{Tables are challenging to extract as they can be presented in a variety of styles and structures. Graphical ruling lines sometimes do not exist (a) and when present (b), may not be a necessary condition to delineate a cell (red box). }
    \label{fig:tough_tables}
\end{figure}

In real world enterprise and scientific applications, crucial information is often summarized in tabular form within PDF or scanned documents~\cite{burdick2020table}.
Since neither of these widely-used document formats preserve logical table structure, accurate table detection and cell structure recognition techniques are required to reconstruct the table before its contents can be leveraged for any subsequent analysis, such as question answering~\cite{Pasupat2015CompositionalSP}, scientific leaderboard construction~\cite{hou2019identification} or knowledge base population~\cite{rotmensch2017learning}. 
Accurate table extraction is possibly the most important task and a major pain point in document analysis for businesses where the computer vision community can have a significant impact.
In fact, the reliance on rules, lack of labelled data and visual nature of table recognition in documents resembles research in the early days of object recognition in images. 
%These make accurate table extraction one of the most important tasks and a major pain point for businesses that has not been significantly tackled by the vision community. 
\emph{Table detection} refers to detecting the boundary of a table, while \emph{cell structure recognition} generates the logical relations of cells and their contents inside a table, e.g., identification of all cells within the same row or column inside the table.
Although straightforward for humans, accurately reconstructing table boundary and cell structure information from PDF or image documents is difficult for automated systems due to the wide variety of styles, layout and content tables have across heterogeneous document sources ~\cite{Hoffswell2019InteractiveRO}. Such visual ``clues'' often conflict across sources, e.g., examples in Figure~\ref{fig:tough_tables}.

Unfortunately, conventional rule-based or statistical techniques for table extraction often fail to generalize as they rely heavily on hand-crafted features like graphical lines or bold font, which are not robust to style variations across different document formats.
Compared to these approaches, vision-based deep learning methods have two advantages. First, by working directly on images, they can be applied to any document renderable to an image, including PDF. They do not rely on programmatic PDF encodings such as graphical line, spacing and font attributes which rule-based approaches require. 
Second, if a large annotated dataset for tables is available, models can be pretrained and then finetuned using a small amount of in-domain labels.
However, few vision-based deep learning models for table extraction that have been proposed, with most existing deep learning approaches directly use off-the-shelf object detectors~\cite{Ren2015FasterRT,Schreiber2017DeepDeSRTDL,Li2019TableBankTB} without any major architectural adaptation.

%As a result, such vision-based deep learning models have not yet surpassed traditional non-deep learning methods in all measures~\cite{Gbel2013ICDAR2T}. 
%We argue the non-ideal results of deep learning methods are due to the lack of large-scale dataset, and over-reliance on off-the-shelf object detectors without adaptation to unique properties of tables.

To tackle cell structure recognition, rule-based and statistical machine learning approaches are commonly used ~\cite{Gbel2013ICDAR2T}. 
Recent deep learning approaches either output structure as text with a image-to-sequence method ~\cite{Vinyals2014ShowAT,Li2019TableBankTB},  or generate structure after detecting related objects in the table.
Although object detection also needs a box to structure conversion step when compared to end-to-end sequence generation, the visualized bounding boxes of object detection methods are easier for humans to interpret and correct, which leads to better results~\cite{Hoffswell2019InteractiveRO}.
Most existing work on object detection-based methods detect entire rows and column separately, and represents the intersection of detected rows and columns as cells~\cite{Schreiber2017DeepDeSRTDL,tensmeyer2019deep}. 
Such an approach has limitations in accurately detecting structure of complex tables with rows or columns which do not span the entire table or align well.
Our proposed Global Table Extractor (GTE) adapts vision-based models to the table identification and cell structure recognition problem, and achieves state-of-the-art results by addressing limitations of existing work as follows.
First, GTE improves object detectors by explicitly enforcing the model to learn the natural constraint of tables: A table must contain certain amount of cells inside it and a cell cannot exist outside of the table.
In other words, the model should not only focus on the tables, but also pay attention to the cells inside. 
Second, we propose to detect each cell directly instead of detecting entire rows and columns separately since cells are more visually distinct as object units and this approach naturally supports tables with rows and columns not spanning the entire table.
Third, current object detection models focus on the local area around objects, which neglects the global style of tables that determine cell appearance. 
To leverage the information of the whole table, we propose a hierarchical system of networks where we discriminate the global context first, the table style. 
%The style model first classifies if a table contains graphical ruling lines.
The table image is then fed into different object detectors specialized for different styles.
After cell bounding boxes are detected, we invent a cell cluster-based algorithm to generate cell structures. 
In summary, our contributions are as follows:\vspace{-0.12cm}
\begin{enumerate}
    \item We present our systematic framework for vision-guided joint table detection and cell structure recognition, GTE, which outperforms previous systems on the ICDAR 2013 and 2019 table competition benchmark.\vspace{-0.1cm} %We also show more than 45\% accuracy improvements testing on an out-of-domain dataset when compared to naive object detection networks trained with but not adapted for tables. We are currently undergoing an internal process to open source the code. 
    \begin{enumerate}
        \item We leverage a cell detection network to guide the training of the table detection network.
        \item We present a hierarchical network and a novel cluster-based algorithm for cell structure recognition by classifying tables, detecting cells and convert this into structure with spatial clustering.
    \end{enumerate}
    \item We design a method to automatically create
    ground-truth labels for table recognition and use it to enhance PubTabNet\cite{zhong2019image} and create FinTabNet, which are large datasets from real-world data sources with fine-grained cell structure annotation for table related tasks. PubTabNet enhancements are now available and we intend to release FinTabNet publicly (subject to legal evaluations) to address the lack of such labelled data.%, which is useful for training deep-learning based models.      
\end{enumerate}

\section{Related Work}

\subsection{Table Detection}
Rule-based methods were among the earliest proposed approaches for locating tables inside a document~\cite{Green1995RecognitionOT,Hirayama1995AMF,Hu1999MediumindependentTD,Gatos2005AutomaticTD,Shafait2010TableDI}. 
Such rules mainly focus on text-block arrangement, horizontal and vertical lines, and item blocks. 
Rule-based systems perform well on some documents, but require extensive human effort to summarize rules and often fail to generalize to other domains or across heterogeneous table formats. 
Statistical machine learning approaches have been proposed to fill these gaps. Unsupervised methods use bottom-up clustering of word segments~\cite{Kieninger1998TheTT}. 
Examples of supervised methods include learning a MXY tree to represent a table~\cite{Cesarini2002TrainableTL}, learning a Hidden Markov Model designed for table structure~\cite{Silva:2009:LRH:1634930.1635458} and learning a SVM to classify tables using line information~\cite{6628801}.
Semi-supervised methods have also been proposed to leverage unlabelled documents~\cite{FanK15}. Recently, data-driven vision based approaches have been used to detect tables by adapting state-of-the-art object detectors such as Faster-RCNN to table detection~\cite{Schreiber2017DeepDeSRTDL,Li2019TableBankTB,Gilani2017TableDU}. 
\subsection{Cell Structure Recognition}
Earliest successful system is the rule-based T-RECS by evaluating horizontal and vertical structure of words~\cite{Kieninger1998TheTT}. 
Wang et al. presented a seven-step process similar to the X - Y cut algorithm to improve the previous system with statistical learning approaches from a training corpus~\cite{Wang2004TableSU}.
Shigarov et al. decomposed tables by offering configuration of algorithms, thresholds and rule sets based on PDF metadata~\cite{Shigarov2016ConfigurableTS}. 
Recently, there is a trend from rule-based and statistical machine learning to deep learning methods in table recognition. Deep learning approaches include two categories: (a) End-to-end image-to-sequence models ~\cite{Li2019TableBankTB, zhong2019image}; (b) Object detection based methods~\cite{Schreiber2017DeepDeSRTDL,tensmeyer2019deep, prasad2020cascadetabnet}.

\subsection{Existing Datasets}
\label{sec:existing-datasets}
During the development of GTE, we found few existing datasets with any kind of structure annotation. We required a dataset with a large number of labelled examples where each table cell is annotated with its pixel-coordinate location, logical coordinates inside the table structure (e.g., row-span and col-span) and cell text contents. Although the ICDAR2013 dataset met the annotation requirements, only 254 table examples (96 train and 156 test from the competition) were available, which were from European Union and US Government reports~\cite{Gbel2013ICDAR2T}. TableBank has 145K labelled tables, but provides only logical coordinates of cells in the table \cite{Li2019TableBankTB}.  
While in the enhanced PubTabNet and FinTabNet dataset, annotations give detailed information on the logical structure as well as the location and contents of each cell, similar to the ICDAR2013 competition.
Very recently, a new ICDAR2019 table competition was held with not PDF files but images of document pages~\cite{gao2019icdar}.
It contains in total 80 documents for table structure recognition, including both modern and handwritten archival documents. 
They do not have a training set for modern documents, only some for testing. 
Other existing datasets only contain table boundary information \cite{Shahab2010AnOA,Siegel2018ExtractingSF}.

\section{PubTabNet, FinTabNet}
\label{sec:sd-dataset}
As shown above, there is a lack of large scale datasets for cell structure recognition. 
To fill this gap, we designed a novel method to automatically match PDF and HTML documents in order to generate a large and comprehensive table recognition dataset. We collaborated with the authors of PubTabNet to enhance the dataset with cell labels, which was originally sourced from PubMed scientific articles. We also worked with them to make a subset of PubLayNet and PubTabNet such that each page has full table and cell information, which we call PubXNet.
To generate the cell structure labels, we use token matching between the PDF and HTML version of each article. 
From the HTML, we know the logical structure of the table cells and from the PDF, we know the cell and table boundary location. 
%There are sometimes token matching errors due to the same tokens in multiple locations or token encoding differences so we also manually verify and correct some articles in the dataset. 
PubTabNet contains more than 568k tables and PubXNet contains more than 24K pages. 

On top of enhancing PubTabNet, we also created FinTabNet, which is a large dataset containing complex tables from the annual reports of the S\&P 500 companies. Financial tables often have very different styles when compared to ones in scientific and government documents, with fewer graphical lines and larger gaps within each table and more colour variations. There are more than 70K pages with full table bounding box and structure annotations (train/val/test= 61801/7191/7085) and more than 110k tables with cell bounding boxes (train/val/test= 91596/10635/10656). The test and validation split are retrieved at the company level with 50 companies in each and companies were selected to have a similar number of tables such that the test sets are not biased towards a particular company.

\section{Methods}
\begin{figure*}[h]
    \centering
    \includegraphics[width=1\textwidth]{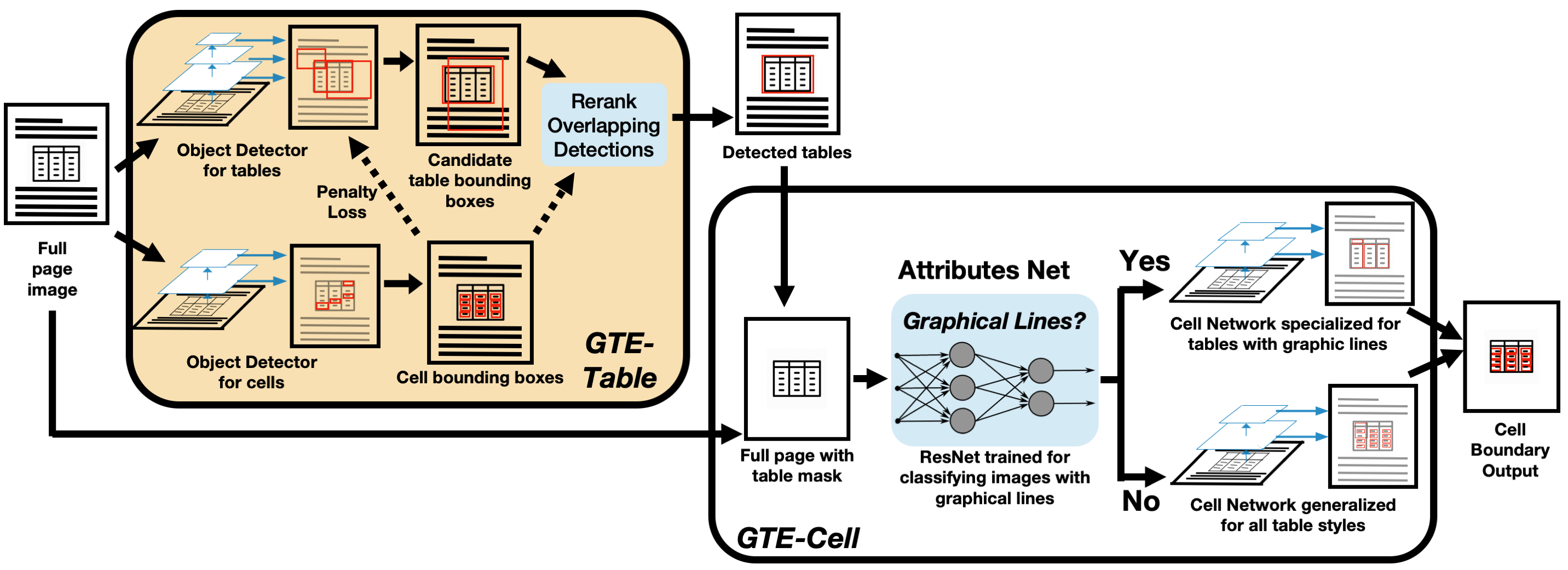}
    \caption{Our full GTE Framework consists of several networks for table (GTE-Table) and cell (GTE-Cell) boundary detection. The input is an image form of a document page for both sub-frameworks, but note GTE-Cell depends on table boundaries output by GTE-Table to generate cell structures for each specific table.}
    \label{fig:gte_framework}
\end{figure*}
As shown in Figure \ref{fig:gte_framework}, our full GTE framework consists of a series of vision-based neural networks. 
Each of the main object detection networks use context from the output of the other networks. 
The framework could be adapted to any kind of object detector.
The table boundary network (GTE-Table) uses a cell detection network by leveraging the fact that tables must contain at least some cells. 
The cell structure recognition network (GTE-Cell) uses table boundaries from the table boundary network(GTE-Table) and table-level style information (Attributes Net). 

\subsection{GTE-Table}
% We leverage TableBank and \sd\, table boundary to pretrain the object detection network before fine-tuning on the ICDAR train set for the table boundary detection task\cite{Lin2017FocalLF}. 
% %To make the model converge faster, it is necessary initialize the parameters pretrained on MS COCO first\cite{Lin2014MicrosoftCC}. 
% As objects in the wild are at very different scales and aspect ratios when compared to tables on pages, we make a few key parameter changes to the original network. More details are in the supplementary material.
%We create a novel cell constraint based loss function, which could be built on top of any object detectors. 
\begin{figure}[htbp!]
    \centering
    \includegraphics[width=0.5\textwidth]{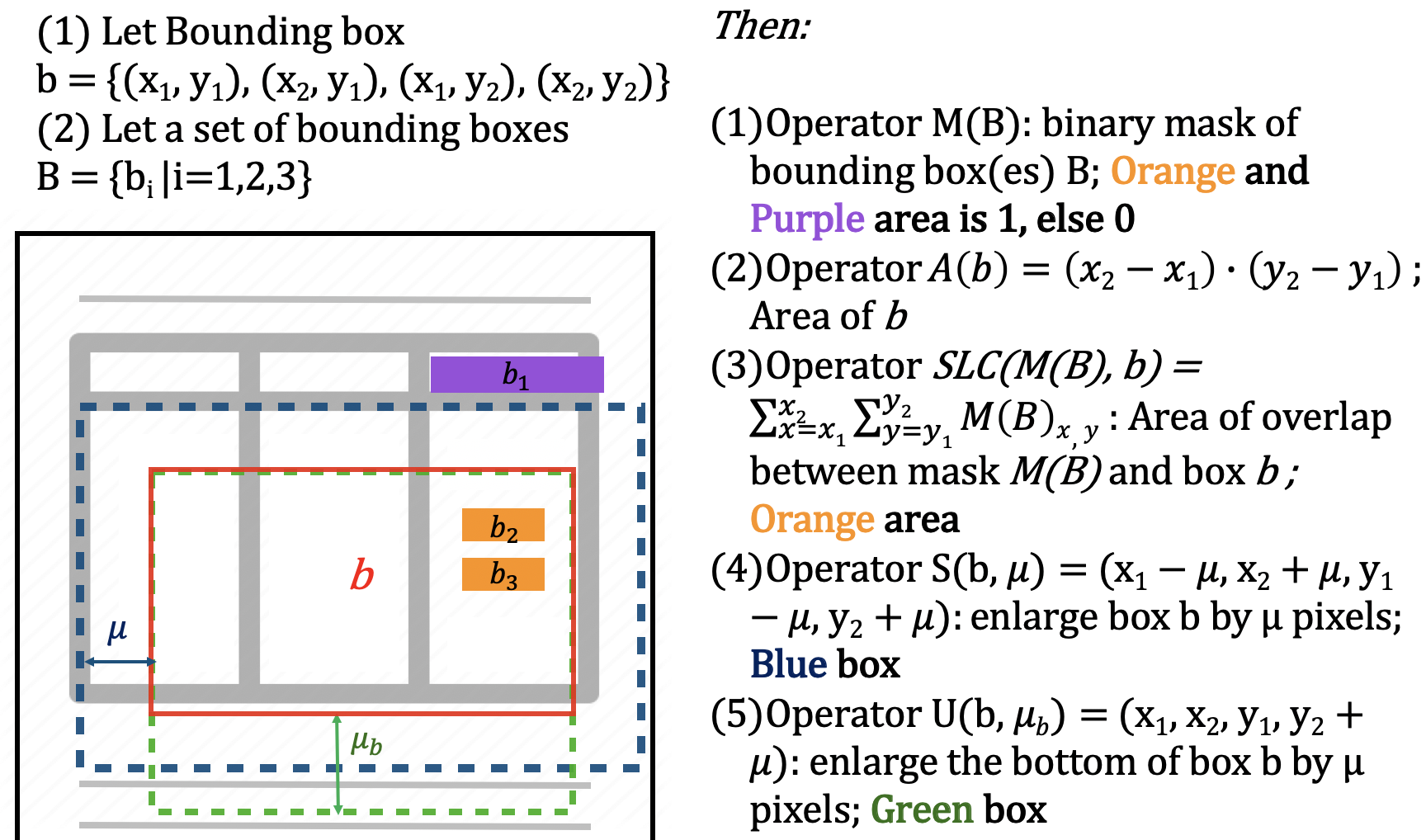}
    \caption{Definition of Operators used in Sec 4.2.}
    \label{fig:loss}
\end{figure}
In the training stage, besides the regression and classification loss, we add a piecewise constraint loss. 
It penalizes the detection probability of unrealistic tables when considering cell locations. 
This novel cell constraint based loss function may be added to any detection network.
We formalize the terminologies of this section here. We make the following definitions in Fig. \ref{fig:loss}.
We used the guided cell network to generate a set of cell bounding box(es) $B_{cells} = \{b_{cell,i}|i\}$. The cells are detected by a simpler non-hierarchical version of our GTE-Cell network that is trained on only original full-page document pages, without knowing the location of the tables.
Given $B_{cells}$, we define two Boolean operators where inputs are an inner box($b_{ibox}$) and outer box($b_{obox}$), which define the boundaries of the mask input area: 
\begin{align*} \label{eq:C}
    C(b_{ibox}, b_{obox}) = & \big\{SLC(M(B_{cells}), b_{obox}) \\
    & - SLC(M(B_{cells}), b_{ibox}) \big\} \\ 
    & < \big\{\alpha \cdot (A(b_{obox}) - A(b_{ibox})) \big\}\\
    D(b_{ibox}, b_{obox}) = & \big\{SLC(M(B_{cells}), b_{obox}) \\
    & - SLC(M(B_{cells}), b_{ibox})\big\} > 0
\end{align*}

Where C is true if the area covered by the cells between $b_{obox}$ and $b_{ibox}$ is at most $\alpha$ times the area of the $b_{obox}$ minus area of the $b_{ibox}$. D is true if any cells exist in the area between $b_{ibox}$ and $b_{obox}$.  The penalty indicator $I(b_{tbl})$ is defined as
\begin{align*} 
    I(b_{tbl})  = & C((0,0,0,0), b_{tbl}) \vee C(S(b_{tbl},\mu_1), b_{tbl}) \vee\\
    & D(S(b_{tbl}, \mu_2), S(b_{tbl}, \mu_3)) \vee
    C(U(b_{tbl}, \mu_4), b_{tbl})
\end{align*}
The penalty indicator is true when any of the following conditions are true:
\begin{itemize}
    \item $C((0,0,0,0), b_{tbl})$ : Less than $\alpha$ of the whole table has cells.
    \item $C(S(b_{tbl},\mu_1), b_{tbl})$ : Less than $\alpha$ of the area just inside the table has cells.
    \item $D(S(b_{tbl}, \mu_2), S(b_{tbl}, \mu_3))$: The area just outside of the table contains any cells. 
    \item $C(U(b_{tbl}, \mu_4), b_{tbl})$: Less than $\alpha$ of the area at just inside the bottom of the table has cells.
\end{itemize}
Then the constraint loss($CL$) is\vspace{-0.2cm}
\begin{equation}\label{eq:cl}
    \sum\limits_{b_{tbl}}^{B_{tbl}} I(b_{tbl})P(b_{tbl}) + \gamma_1 (1-I(b_{tbl}))(1-P(b_{tbl}))
\end{equation}
where $P(\cdot)$ is the table detection probability function. 
%We choose $\mu_x=0.8\%, \mu_y=1\%, \mu_b=1\%, \alpha = 0.25, \gamma=0.1$ in our experiments.
We choose $\mu_1 = -5$, $\mu_2 = 5$, $\mu_3 = 10$, $\mu_4 = -10$, $\alpha = 1/8$, $\gamma_1=1/10$ in our experiments.
Additionally, one of the input image channels to the table network is replaced with a mask generated from the prediction of cells to further aid training.

In the inference stage, instead of the widely used non-max suppression, our ranking of proposed bounding boxes not only consider detection probabilities, but also the presence of cells inside and outside the table. 
We define \textbf{Constraint Coefficient}(CCoef) for each bounding box, where\,$CCoef(b_{tbl}) = 
SLC(M(B_{cell}),S(b_{tbl}, \mu_5)) - SLC(M(B_{cell}),b_{tbl}) - 
\gamma_2\cdot (SLC(M(B_{cell}),b_{tbl}) - 
SLC(M(B_{cell}),S(b_{tbl}, \mu_6)))$. 
For each boundary of the table bounding box, we calculate the amount of cells just outside subtracted by the amount of cells just inside the table.
For any pair of bounding boxes $b_i, b_j$ overlapped with each other more than $\delta\%$, and $|P(b_i) - P(b_j)| < \epsilon \stepcounter{equation}$, we discard the bounding box with higher $CCoef$. We choose $\mu_5 = -20$, $\mu_6 = \{0.25*(x2-x1), 0.25*(y2-y1)\}$, $\gamma_2 = 0.1$, $\epsilon=0.1, \delta=25$ in our experiments. The hyper-parameters are described in more details in Supplemental material. Here it suffices to say that they are chosen, in a straightforward way, based on characteristics of tables in typical documents; to give some intuition of the concrete values, in the above, a value of 5 reflects half of the height of a character (10 pixels), while 20 corresponds to two lines of text. 

\begin{figure*}[htbp!]
\centering
    \begin{subfigure}[b]{0.22\textwidth}
        \includegraphics[width=\textwidth]{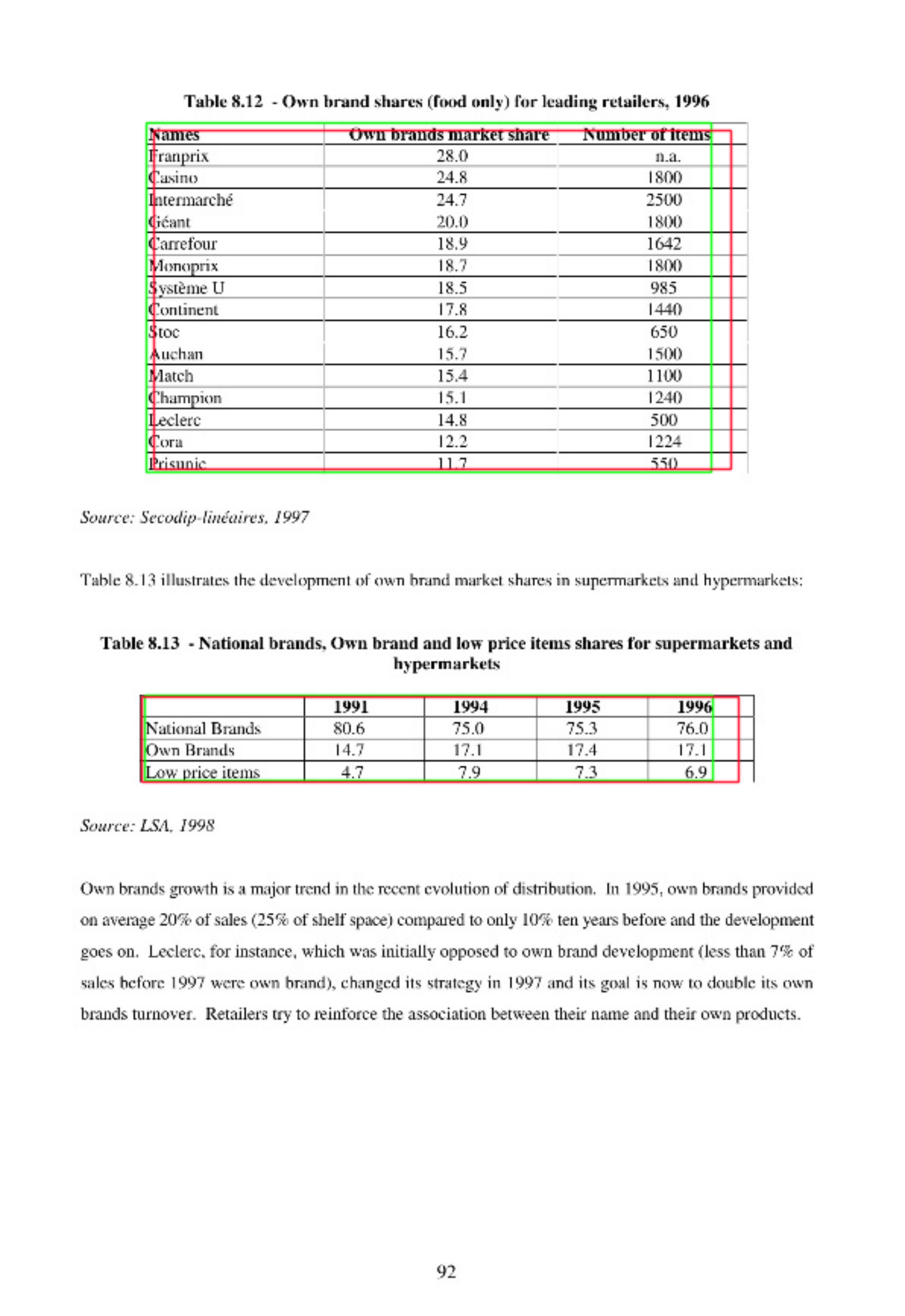}
        \caption{}
    \end{subfigure}
    \begin{subfigure}[b]{0.22\textwidth}
        \includegraphics[width=\textwidth]{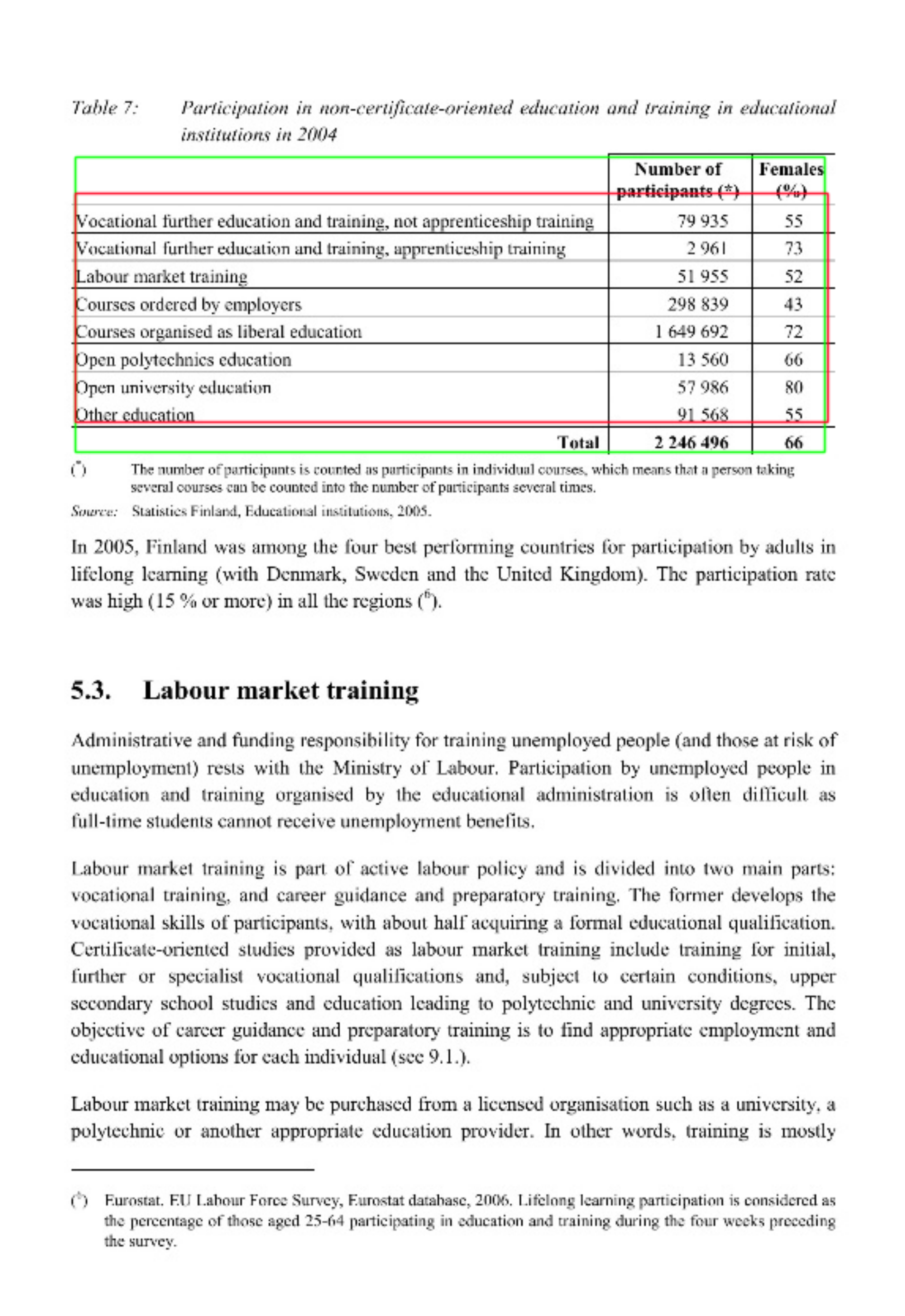}
        \caption{}
    \end{subfigure}
    \begin{subfigure}[b]{0.22\textwidth}
        \includegraphics[width=\textwidth]{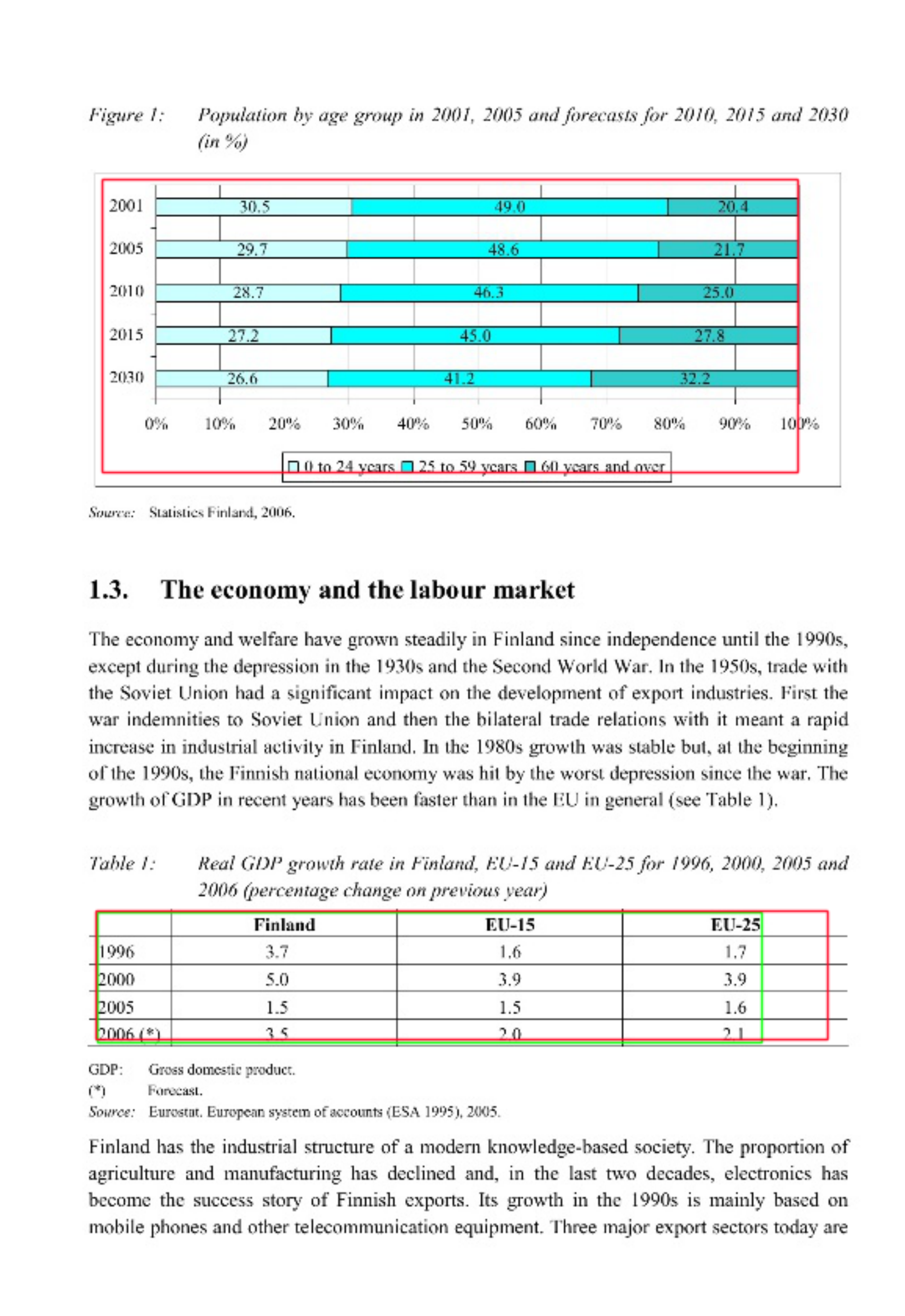}
        \caption{}
    \end{subfigure}
        \begin{subfigure}[b]{0.22\textwidth}
        \includegraphics[width=\textwidth]{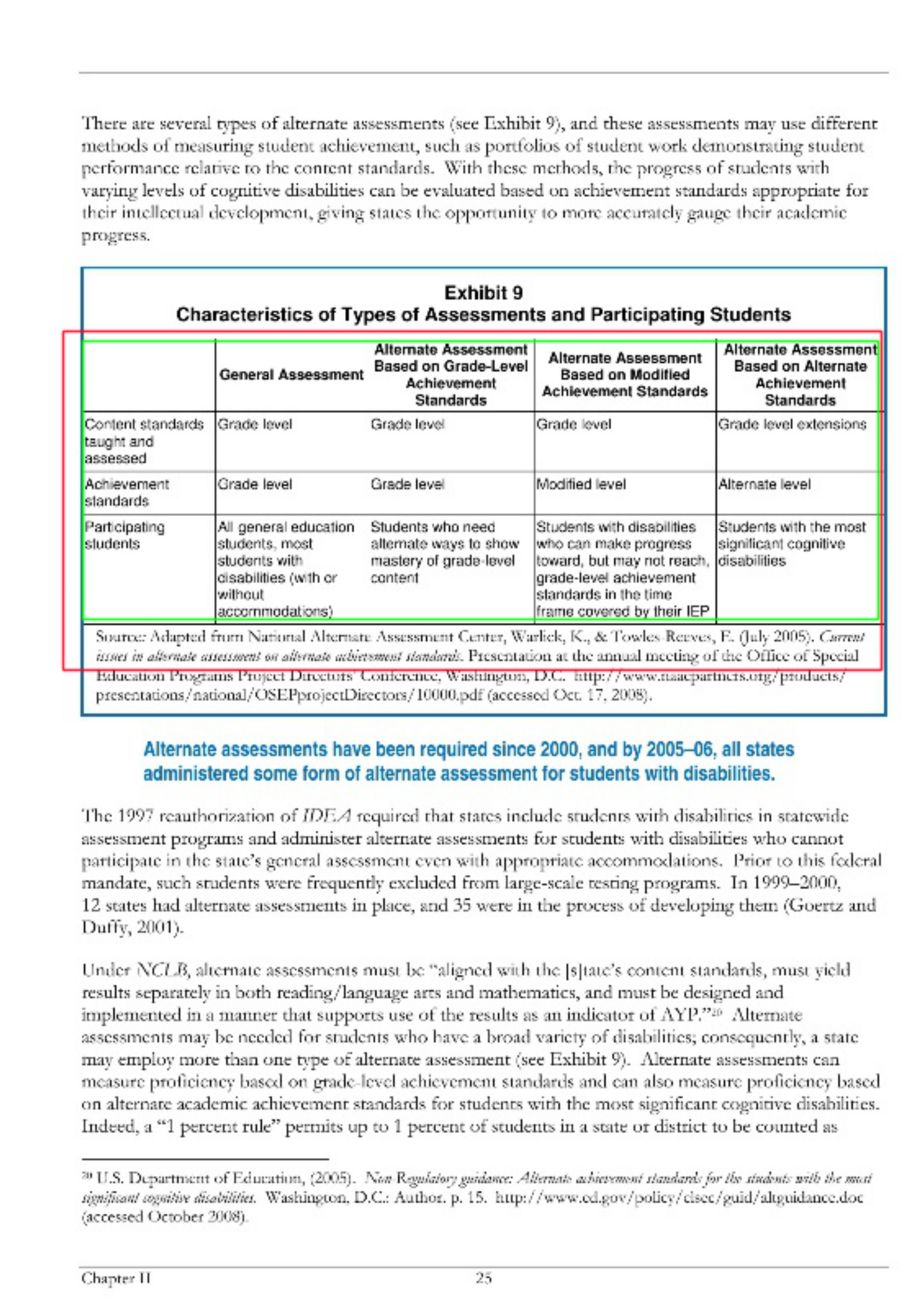}
        \caption{}
    \end{subfigure}

    \caption{(a) Correct detection\, (b) Partial under-detection\, (c) Mis-detection\, (d) Over-detection}
    \label{fig:table_ex}
\end{figure*}
\subsection{GTE-Cell}

Tables in the real world often adhere to a global style that determines the rules and meanings of its components. 
For example, there are some tables that have visible vertical and horizontal ruling lines for every row and column, easily defining cell boundaries. 
%In this case, it is trivial to say that each cell is within visible boundaries on every side. 
However, there are other styles that have no ruling lines or only intermittent breaks. 
In such a case, a model that only looks at its local surroundings, such as most object detection networks, would not be able to ascertain whether a ruling line represents the start of a new cell.
Empirically, we also found that mixing different styles of tables in training worsens model performance on some data, even though this used more training data. 
In our framework, we first train an attribute network aimed at classifying the presence of vertical graphical ruling lines in the table.
The output of this network determines which of two cell detection networks is used, which were trained with different augmentation schemes. 
The ``no lines" scheme erases all existing graphic lines and ``full boundaries" adds in vertical and horizontal boundaries for every row and column at the median point between cells. 
The network specialized on tables with graphical lines is trained on the original plus ``full boundaries" data while the other network is trained on the original and all augmentations. 
%The final output of the network generates bounding boxes for each cell. 

To convert the bounding box output into a logical structure, we first align cell boxes to text lines as extracted from the PDF. 
Then, we determine the number of rows and columns by sampling in the vertical and horizontal directions, respectively. 
Before sampling in the vertical direction to determine the number of rows, we expand the left and right edges of cells while it is not overlapping with existing cells, to account for rows with missing cells. If there are graphical lines available, we ensure that the number of rows or columns detected are at least equal to the number of unique inner lines plus one. 
Then, we infer the vertical and horizontal alignments of the table by which edge of the cell box has the best alignment with other cells. 
We use K-means clustering on cell bounding box coordinates to define row and column locations. 
Then, we assign row and column positions to each cell based on their box locations, merging cells when necessary. 
Finally, we leverage the fact that cell content generally starts with a capital letter. Therefore, cells that start with a small case is likely a case of over-splitting. 
We merge these cells with the cell above. 
Also, we perform some post-processing steps. This includes assigning locations to leftover text boxes that were not overlapping with any detected cells and we split cells in certain cases when there are gaps nearby. 
Before producing the final logical structure of each cell in the table, we increase the row and column span of cells when the text box intersects with neighboring empty rows or columns as this is likely a hierarchical cell spanning multiple rows or columns. 
Our clustering-based algorithm is more efficient than a greedy or exhaustive search method that selects each cell sequentially. 
As well, many of our steps are designed to be robust against cell detection errors.
%and the span does not cause two cells to occupy the same space. 
For more details, see Algorithm 1 in the supplementary material. 

\begin{table*}[htbp!]
\caption{Table Detection Percent Results on ICDAR2013. We also provide purity and completeness scores when available. There are a few other methods\cite{siddiqui2018decnt,huang2019yolo,Kavasidis2018ASC} that could not be compared directly in this table as they are using a measure based on Intersection-over-union(IOU) where the IOU threshold=0.5. Our method achieves F1=0.997 by this measure, which is higher than reported by the other methods. We observe the character-based measure computed by the competition script better measures table quality than a measure based on IOU threshold of 0.5, since the latter counts as correct for predictions capturing only half of a ground-truth table which have little practical use. Also, \cite{prasad2020cascadetabnet, Schreiber2017DeepDeSRTDL} used different train/test split from the original competition without publishing their split and so cannot be compared directly. For brevity, we present only the highest performing method in each category. The full table is in the supplement}
\centering
\scalebox{0.9}{
	\begin{tabular}{c|c|c||ccc|cc}
	    \hline
	    Category & Method & Input type & Recall & Precision & F1 & Cpt & Pu \\
	    \hline \hline
	    {Commercial Softwares} & \textsl{FineReader} & PDF & 99.71 & 97.29 & 98.48 & 142 & 148\\\hline
	    {Non Deep Learning} 

	    & \textsl{Nurminen}\cite{Gbel2013ICDAR2T} &  PDF & 90.77 & 92.10 & 91.43  & 114 & 151 \\\hline

	    {Deep Learning} 
    & \textsl{TableBank}\cite{Li2019TableBankTB}   & Image & / & / & 96.25  & / & /\\\hline
        Ours & GTE & Image & \bf{99.77} & \bf{98.97} & \bf{99.31} & 147 & 146\\
	    \hline \hline
	    \multirow{2}{*}{Ablation}  &   Detection-Base & Image & {84.64}& {90.65} & {84.65} & {68} & {97} \\
	    &GTE-Table-Sep & Image & 95.71 & 98.18 & 95.71 & 140 & 150 \\\hline\hline
	    %GTE-Table-Joint & 97.73 & 95.90 & 96.24 & 141 & 145 \\\hline
	   %& GTE-Table-Joint & Image &  \bf{97.77} & \bf{98.97} & \bf{99.31} & 147 & 146 \\\hline

	\end{tabular}
	  	}
  	\label{tab:td_results}
\end{table*}

\section{Experiments}

\subsection{Datasets}
We perform extensive experiments on both the table detection and cell structure recognition tasks in the widely used ICDAR2013 table competition \cite{Gbel2013ICDAR2T}. 
This dataset is considered as a standard benchmark dataset in PDF table extraction. 
It contains 96/156 tables for training/testing collected from European Union and US Government reports.
Since the in-domain dataset is very small, pretraining the model on other datasets is required.
For table detection, we pretrain the model on the combination of TableBank(\cite{Li2019TableBankTB}) and PubTabNet; For cell structure recognition, we pretrain the model on PubTabNet.

We also conduct additional experiments on ICDAR2019 as well as the PubTabNet and FinTabNet datasets. 
%As our model is not trained on financial filing data, it can be considered out-of-domain results. 
\subsection{Evaluation Metrics}
For ICDAR2013, We use the official evaluation script of ICDAR2013 table competition \cite{Gbel2013ICDAR2T}. 
For ICDAR 2013 table detection, the metrics are character-level Recall (Rec.), Precision (Prec.) and F1-measure op(F1), averaged per document, along with Purity (Pu) and Completeness (Cpt). Consider $N$ is the set of test documents, then they are defined as follows:\vspace{-0.3cm}
$$Pu  = \sum\limits_{n \in N} \lfloor Rec(n) \rfloor \quad Cpt = \sum\limits_{n \in N} \lfloor Prec(n) \rfloor \vspace{-0.3cm} $$

For cell structure recognition, the metrics are precision, recall and F1-measure for generated adjacency matrices. Additional details are available in \cite{Gbel2012AMF} and \cite{Gbel2013ICDAR2T}.

\subsection{Experimental Setup}
\label{para:setup}
\subsubsection{Training and Inference Details} We leverage TableBank and PubTabNet table boundary to pretrain the object detection network before fine-tuning on the ICDAR train set for the table boundary detection task\cite{Lin2017FocalLF}.
We use the architecture of RetinaNet with Resnet50-FPN backbone as our base object detection model \cite{Lin2017FocalLF,Lin2016FeaturePN}. 
We use resolution of 643 by 900 for tables, and 965 by 1350 for cells, as cells need higher resolutions to distinguish. 
We redesigned the feature pyramid network for tables and cells such that there are fewer detection layers than a typical object detection network but this allows for finer-grained anchor boxes for cells and larger object boxes for tables without sacrificing computational efficiency. 
%We only consider feature pyramid $P3$ and $P5$ for cells, and $P4$, $P5$, $P7$ for tables. We adjust the choice of feature maps based on the distribution of cell and table sizes in train set. 
We add anchors with aspect ratio 0.1 and 0.25 for each feature map to catch commonly appearing wide tables and cells. 
In the cell network, since the objects are really dense, we use anchors of sizes {0.5, 0.7, 1, 1.2, 1.6} of the set of aspect ratio anchors. 
We add additional smaller scale anchors because many cells are much smaller than the anchors.
In the table network, we run each page at test time at multiple zoom scales to help improve detection of abnormally small or large tables. %These are at 95\%, 90\%, 105\% and 110\%. 
All the object detection models in GTE are initialized with the parameters pretrained on MS COCO dataset  \cite{Lin2014MicrosoftCC}. 

\subsection{Experimental Results}
\subsubsection{Table Detection}
As reported in Table \ref{tab:td_results}, GTE-Table achieves the best character-level F1 measure among all methods. Although \textsl{FineReader} slightly outperforms GTE on purity, the higher F1-measure for GTE indicates GTE produces higher quality boundaries closer to ground-truth.  Since the purity metric penalizes all incorrect table boundaries equally, it does not provide ``partial-credit'' for almost correct answers in the same manner as character F1-measure for cases where the predicted boundary only includes a few extra characters. 
Figure \ref{fig:table_ex} shows some correctly detected table boundaries as well as some failures. In general, we see three types of errors, partial under-detection, where some parts of the ground truth table is missing, partial over-detection, where some text outside of the ground truth is mistakenly included and mis-detection, where a non-table entity such as a chart was misidentified as a table. We do not see any cases of table non-detection in our ICDAR2013 test results and only one case of mis-detection. Overall, most partial detections are only missing or adding one or two extra lines, such as a short captions in the table.

\subsubsection{Table Detection Ablation Study}
% \begin{table}[htbp!]
% \caption{Table detection additional experiments with percent results. }
%     \centering
%     \begin{tabular}{c||ccc|cc}
% 	    \hline
% 	    Method & Rec. & Prec. & F1 & Cpt & Pu \\
% 	    \hline \hline
% 	    Detection-Base & {84.64}& {90.65} & {84.65} & {68} & {97} \\\hline
% 	    GTE-Table-Sep & 95.71 & 98.18 & 95.71 & 140 & 150 \\\hline
% 	    %GTE-Table-Joint & 97.73 & 95.90 & 96.24 & 141 & 145 \\\hline
% 	    GTE-Table-Joint & \bf{97.77} & \bf{98.97} & \bf{99.31} & 147 & 146 \\\hline
% 	    %GTE-Table-Joint  & \multirow{3}{*}{\bf{99.77}} &  \multirow{3}{*}{\bf{98.97}} &  \multirow{3}{*}{\bf{99.31}} & \multirow{3}{*}{147}  & \multirow{3}{*}{146} \\
% 	    %w/ reranking  & & & & \\
% 	    %and multi-zoom & & & & \\
%     	    \hline

% 	    \hline
% 	\end{tabular}
     %GTE-Cell-Table-Comb is the multi-task network performing cell and table detection. GTE-Table-Sep is a dedicated network for table detection but does not take into account any cell network outputs, unlike GTE-Table-Joint. GTE-Table-Joint w/reranking not only includes the additional cell based loss function, it also reranks overlapping outputs at test time according to cell content inside and near the table. }
%     \label{tab:table_addl_results}
% \end{table}
\vspace{-0.1in}
As shown by the additional experimental results in Table \ref{tab:td_results}, the base detection network trained to perform the cell and table detection task simultaneously (Detection-base) performs far worse than the more specialized networks. 
There are two main reasons behind this. 
First, TableBank data cannot be leveraged when pretraining the networks because it lacks cell bounding boxes annotations \cite{Li2019TableBankTB} so it is only trained of PubTabNet and finetuned on ICDAR training data. 
Second, tables and cells are of two completely different scales where it is hard to choose an appropriate resolution to generate anchors fitting the two scales.
On the other hand, it is still important for the cell network and table network to leverage each other's information, as shown by the nearly 3\% boost in F1 accuracy as compared to the regular object detection losses (GTE-Table-Sep) that do not use information from other networks.

\subsubsection{Cell Structure Recognition}

% Final cell structure results
\begin{figure}[htbp!]
    \centering
    \includegraphics[width=0.42\textwidth]{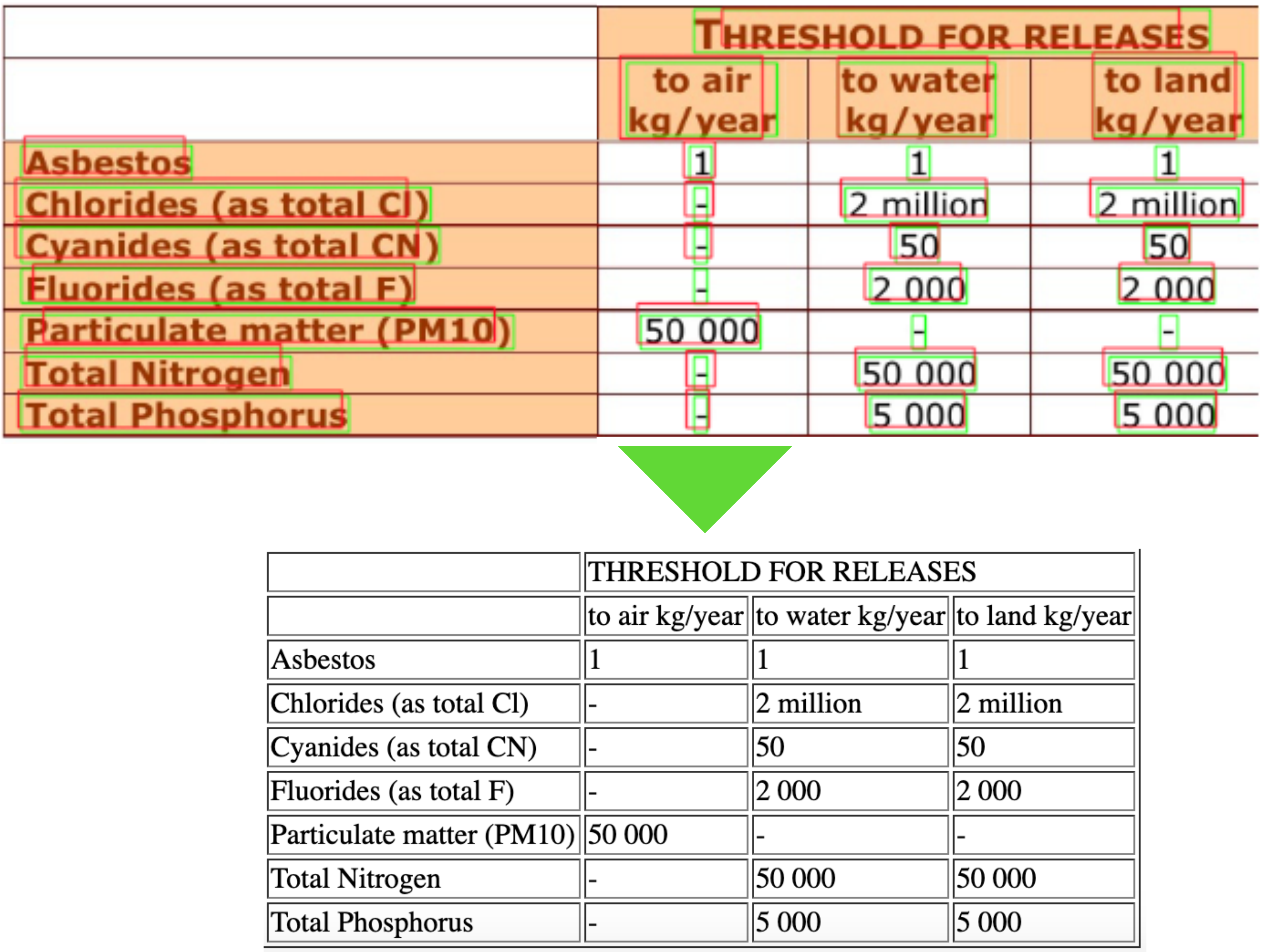}
    % \caption{(a) Long texts partial detection with correct cell structure\,(b) Mis-detected cells with correct cell structure\,(c) Overmerged cells with wrong structure }
    \caption{Partial cell detections with correct cell structure}
    \label{fig:cell_ex}
\end{figure}

As reported in Table \ref{tab:cstr_results}, GTE-Cell outperforms all previous methods and commercial software in all metrics even without using any PDF encodings (ruling lines, rendering techniques, etc). 
All results are from cell detection on outputs produced from table detection by each framework, not the ground truth table. 
When analyzing the qualitative results in Figure~\ref{fig:cell_ex}, we see cell boundary detection often generates a detection box that is too short for very long lines of text. 
This is a key limitation of the anchor-based object detection system, which has difficulties with aspect ratios differing greatly from ones in the configuration. 
As well, in the case of tables without graphical lines at every row and column, the model may mistakenly merge multiple cells into one. 
In many cases, our post-processing boundary to structure algorithm is robust to some of these mistakes are still able to generate a correct or nearly correct structure output. 
We see three main types of detection errors that can lead to incorrect structure output.
There are overmerged cell detection, where two or more cells are incorrectly merged together, oversplit cell detection, where one cell has been incorrectly split into multiple cells and cell non-detection, where there is no predicted bounding box that includes such a cell. 
These errors can lead to a number of inaccuracies in the boundary to cell structure process, including incorrect number of rows and columns, alignment and of course final cell location assignment as well.
Examples of such errors are in the supplementary material. 

\subsubsection{Cell Structure Ablation Study}
\vspace{-0.2cm}
\begin{table}[h!]
    \caption{Cell Structure results on ICDAR2013 show that GTE improves previous state-of-the-art in cases where the ground truth table border (GT?) was and was not used. For brevity, we are only presenting the highest performing method in each category. The full table is in the supplement.}
    %To note, our F1-score is also higher than that of three other related methods, but they could not be compared directly in this table. DeepDeSRT\cite{Schreiber2017DeepDeSRTDL} and DeepTabStr\cite{siddiqui2019deeptabstr} split the ICDAR2013 test set into two parts, leaving only 31 of 156 tables for testing and the rest incorporated into training. To the best of our ability, we have not found which 31 tables were selected. \cite{Shigarov2016ConfigurableTS} presented in the paper that they used a different evaluation script, so it is not directly comparable to other numbers reported in ICDAR2013 table competition. }
    \centering
    \scalebox{0.9}{
    \begin{tabular}{c|c|ccc}
	    \hline
	    Method  & GT? & Rec. & Prec. & F1  \\
	    \hline \hline
	    
	    \textsl{Nurminen}\cite{Gbel2013ICDAR2T} & N & 80.78 & 86.93 & 83.74 
 \\\hline 
        GTE & N & \bf{92.72} & \bf{94.41} & \bf{93.50}  \\\hline \hline
	     \textsl{Tensmeyer}\cite{tensmeyer2019deep} & Y &  94.64 & 95.89 & 95.26 \\
	    \hline
	   GTE & Y &  \bf{95.77} & \bf{96.76} & \bf{96.24} \\
	    \hline
	    \hline
	    	  %  \hline
	   % Method & Rec. & Prec. & F1  \\
	   % \hline \hline
	    Detection-Base & Y & 76.66 & 80.63 & 78.10 \\\hline
	    GTE-Cell-Style-Mix & \multirow{2}{*}{Y} & \multirow{2}{*}{89.78} & \multirow{2}{*}{89.30} & \multirow{2}{*}{89.43} \\
	    -no-pt &&&\\\hline
	    GTE-Cell-Style-Mix & Y & {92.39} & {94.20} & {93.15}\\ \hline
	    GTE-Cell-Border & Y & 91.60 & {93.67} & {92.48}\\ \hline
	    
	   % GTE-Cell-Hierarchical & \bf{95.77} & \bf{96.76} & \bf{96.24} \\\hline
	\end{tabular}}
    \label{tab:cstr_results}
\end{table}

To analyze our GTE-Cell network further, we compare the several variations in Table~\ref{tab:cstr_results} using ground truth table borders. 
Firstly, the baseline detection network (Detection-Base) that performs both cell and table detection has very poor recall and precision. 
For networks specialized for cell detection, we see that the model pretraining on the PubTabNet dataset gives a boost when compared to GTE-Cell-Style-Mix-no-pt. 
We also test each of the sub detection networks (GTE-Cell-Style-Mix for the network trained on all augmentations and GTE-Cell-Border trained on original and graphical line augmentations).  
The full hierarchical model GTE-Cell-Hierarchical performs better than both individual sub-models, showing that it is indeed helpful to first determine the style of the table and then use the model trained on data most similar to it. 
Out of 156 total test tables in ICDAR2013, there are 108 with at least some vertical graphical lines (69.23\%). 
To note, our attributes network (graphical line table classifier) was correct in 123 out of 156 tables (78.84\%). 
The errors generally come from very small tables or tables with vertical graphical lines that only span the header, which is an ambiguity also present in the training data. 
To help mitigate this error, during the row and column sampling step, we keep track of the standard deviation of the sampling points. 
If this value is high, it likely indicates that the cell detection model used was not suitable for the given table as tables tend to have similar number of columns and rows throughout, thus we would then use the alternate cell model.  

% \begin{table}[htbp!]
%     \small
%     \caption{Cell structure additional experiments with percent results using ground truth table borders. }
%     \centering

%     \begin{tabular}{c||ccc}
% 	    \hline
% 	    Method & Rec. & Prec. & F1  \\
% 	    \hline \hline
% 	    Detection-Base & 76.66 & 80.63 & 78.10 \\\hline
% 	    GTE-Cell-Style-Mix-Train-no-pt & 89.78 & 89.30 & 89.43 \\\hline
% 	    GTE-Cell-Style-Mix-Train & 92.39 & 94.20 & 93.15\\\hline
% 	    GTE-Cell-Border-Train & 91.60 & 93.67 & 92.48\\\hline
	    
% 	    GTE-Cell-Hierarchical & \bf{95.77} & \bf{96.76} & \bf{96.24} \\\hline
% 	\end{tabular}

    %GTE-Cell-Table-Comb is the multi-task network trained for both table and cell detection. GTE-Cell-Style-Mix-Train-no-sd has the same style augmentation as GTE-Cell-Style-Mix-Train where tables are augmented with additional or fewer graphical lines but it is not pre-trained on \sd . GTE-Cell-Border-Train is trained on original tables and those augmented with graphical lines. GTE-Cell-Hierarchical is composed of an attribute network that determines which of the GTE-Cell-Border-Train or GTE-Cell-Style-Mix-Train to invoke for each input table. Finally, GTE-Cell-Hierarchical+GT Boundary is tested on tables with ground truth table boundaries and so has a higher accuracy than GTE-Cell-Hierarchical, which used our table detection framework.}
%     \label{tab:cstr_addl_results}
% \end{table}

\subsubsection{Experiments with Additional Datasets}

To demonstrate the robustness of our network on more complex tables and ones outside of the training data domain, we tested the same model on ICDAR2019, PubTabNet, and FinTabNet (Table \ref{tab:addl}). 
For ICDAR2019 table border task (Table \ref{tab:icdar19table}), our score is comparable to the top method. However, it can be difficult to really differentiate as we found that many of our table detections are correct but the annotations themselves are inconsistently including or excluding whitespace. Therefore, we believe that the IOU=0.9 measure is not reliable without determining the amount of text correctly included. We also adapt our model output for task B2 for modern documents (trained on FinTabNet as no training data is provided) to demonstrate our full recognition system and we show significant improvements to both the competition top-performer as well as more recent results(Table \ref{tab:icdar19cell}). We believe that the IOU at lower thresholds is a more accurate measure for this task similar to our reasoning for table border as we found that many of our structure is exactly correct when looking at the text extracted but is shown as 0 at IOU=0.9 as the small text boxes are a bit shifted from the label without cutting off any text so we show results for IOU=0.1 as well. Examples of this ambiguity is displayed in the supplement. 

For PubTabNet and FinTabNet data, we use TEDS scoring (see details in ~\cite{zhong2019image}) to be consistent with the original PubTabNet paper. However, these numbers are not directly comparable as our results are on the PDFs from the validation set. The test set has not been made available. Additionally, we noticed that the original dataset inconsistently included bolding and italics that are not in the original table image, we therefore modified the original evaluation script to ignore these styling tags. 
Nevertheless, our TEDS of 93.01 compares well to the original score of 88.38.
We also show good table and structure recognition scores on our new FinTabNet dataset both in cases where the model was and was not finetuned. It performs much better than the detection-base, showing that our model improvements transfer to other document domains. 
%We show that although these tables are more complex as they rarely have any graphical separation lines, our model is still able to detect the tables and its structure with high accuracy. 
%Additionally, although financial filings were not part of the training set, we still demonstrate good performance on these out-of-domain documents. 
%In particular, there is at least a 30\% improvement in cell structure recognition when compared to the base object detection model. 
%Finally, one may note that there is not a large difference between GTE-Table and Detection-Base for PubTabNet, likely due to the uniform size and look of tables from scientific articles. 
%However, there is still a significant difference when it comes to the cell structure.
%This shows that both GTE-Table and GTE-Cell play important roles in our state-of-the-art table structure recognition performance. 

\begin{table}[htbp!]
    \small
    \caption{Table detection and structure results on scientific paper PubTabNet (PTN) and out-of-domain financial filings FinTabNet (FTN) before and after finetuning (FT?).}
    \centering
\scalebox{0.9}{
    \begin{tabular}{c|c|c|c|c|c}\hline
         Dataset  &  Method & Task & FT? &  Table F1  & TEDS \\\hline \hline
         {\centering PTN} & \textsl{GTE}  & Structure & Y & NA & 93.01 \\\hline \hline
         {\centering FTN} & \textsl{Det-Base}  & Table & N & 81.17 & NA \\\hline
         {\centering FTN} & \textsl{GTE}  & Table & N & 89.97 & NA \\\hline
         {\centering FTN} & \textsl{GTE}  & Table & Y & \bf{95.29} & NA\\\hline \hline
         {\centering FTN} & \textsl{Det-Base}  & Structure & N & NA & 41.57 \\\hline
         {\centering FTN} & \textsl{GTE}  & Structure & N & NA & 87.14 \\\hline
         {\centering FTN} & \textsl{GTE}  & Structure & Y & NA & \bf{91.02}\\\hline
         
    \end{tabular}}
    \label{tab:addl}
\end{table}

\begin{table}[htbp!]
    \caption{Table detection results ICDAR 2019 competition.}
    \centering
    \scalebox{0.9}{
    \begin{tabular}{c|c|c|c|c|c}\hline
         \multirow{2}{*}{Method} &  \multicolumn{2}{|c|}{IOU = 0.8}  & \multicolumn{2}{|c|}{IOU = 0.9} & \multirow{2}{*}{Weighted F1} \\
         \cline{2-5}
         &P&R&P&R\\\hline
          \textsl{NLPR-PAL}\cite{gao2019icdar}  & 93 & 93 & 86 & 86 & 93\\\hline
          \textsl{TableRadar}\cite{gao2019icdar}  & 95 & 94 &\bf{90} & \bf{89} & \bf{94}\\\hline
          \textsl{GTE}  & \bf{96} & \bf{95} & \bf{90} & \bf{89} & \bf{94}\\\hline
         
    \end{tabular}}

    \label{tab:icdar19table}
\end{table}
\begin{table}[htbp!]
    \caption{Cell structure results for ICDAR 2019 competition Task B2-Modern.}
    \centering
    \scalebox{0.9}{
    \begin{tabular}{c|c|c|c|c}\hline
         \multirow{2}{*}{Method} &  \multicolumn{3}{|c|}{IOU}  & \multirow{2}{*}{Weighted F1} \\
         \cline{2-4}
         &0.1&0.5&0.6\\\hline

          \textsl{NLPR-PAL}\cite{gao2019icdar}  &- & 36.5 & 30.5 & 20.6\\\hline
          \textsl{CascadeTabNet}\cite{prasad2020cascadetabnet}  &-& 43.8 & 35.4 & 23.2 \\\hline
          \textsl{GTE}  & 77.5 & \bf{54.8} & \bf{38.5} & \bf{24.8}\\\hline
         
    \end{tabular}}
    \label{tab:icdar19cell}
\end{table}

\section{Conclusion and Future Work}
In summary, we have demonstrated a vision based table extraction framework with state-of-the-art results. 
It can perform the full pipeline of table recognition, from document to table structure, which can be used easily for down-stream analysis. 
Our framework leverages the global visual context of tables, including the style and rules in the relationship between cells and tables. 
As well, we have released the enhanced PubTabNet dataset and will release FinTabNet, which we hope will help others using data hungry methods to tackle table-related problems. 
Our vision based method is very easily merged with Optical Character Recognition (OCR) methods to perform table recognition fully from images. 
%Indeed, in future work, we are adapting the algorithm in this direction. 

\clearpage

{\small
\bibliographystyle{ieee}
\bibliography{main}
}

\end{document}

\section{Visualization of different structure format}
 \begin{figure}[h!]
     \centering
     \includegraphics[width=\textwidth]{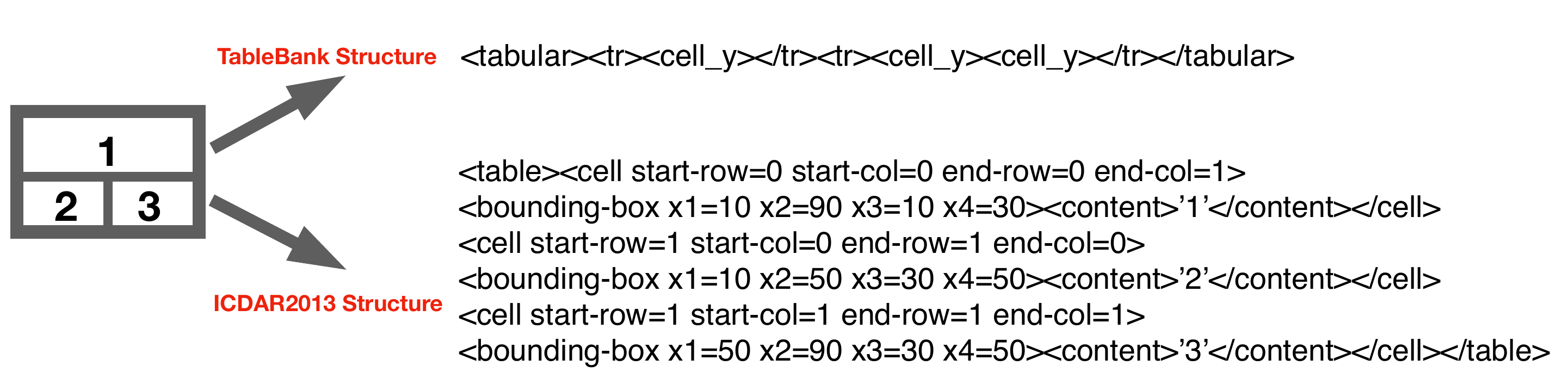}
     \caption{TableBank versus ICDAR2013 structure annotations}
     \label{fig:tb_vs_sd}
\end{figure}

\section{Experimental Details}
\subsection{GTE-Table Network}
We make a few changes to the original RetinaNet model in GTE-Table.  We add anchors with aspect ratio $\{0.1, 0.25\}$ to each feature map for wide tables. The input image size is $900 * 643$. 
\subsection{GTE-Cell Network}
The GTE-Cell Network is composed of a line classifier network at the top of the hierarchy and two object detection models that specialize on different styles of tables.
The graphic line classifier network is a ResNet50 model with a binary classifier on top.
This network is first pretrained with the attributes derived from \sd\,dataset and then fine-tuned on the ICDAR train dataset. 
The ground truth data is derived from the presence of nearby vertical graphical lines (as detected by a PDF parser) for each cell. 
We make the following changes to the original RetinaNet model in GTE-Cell for cell object detection. 
Since the scale of cell is generally small, we use pyramid levels $P3$ and $P5$. 
We find that skipping $P4$ allows us to add additional anchors while keeping a similar level of computational efficiency. 
We add anchors with aspect ratio $\{0.1, 0.25\}$ to each feature map to better detect very wide cells. 
% what does denser scale mean here?
For denser scale objects, at each level we use anchors of sizes $\{0.5, 0.7, 1, 1.2, 1.6\}$ of the set of aspect ratio anchors.
We add additional smaller scale anchors because the majority of cells are much smaller than the anchors generated from $P3$. 
The input image size is $965 * 1350$.

\subsection{Hyper-parameter Selection} For joint training, our hyper-parameters are selected from characteristics of the ICDAR training data. On average, the height of a character is 10 pixels. We wanted to check the text density of tables just inside and just outside of the table; we chose 5 pixels (or half a character height) for this purpose. As a result, we chose $\mu_1 = 5$ and $\mu_2 = 5$. 
We also chose $\alpha = 1/8$ (the density threshold) as we calculated the cell density of tables in the training set and found that the value at the lower end of the density scale (5th percentile) was around 1/8. 
We did not select the minimum (which was around 0.1) in case there are outliers in the training set. 
Finally, $\gamma_1 = 1/10$ in Eq.\ref{eq:cl}  gives less penalty to false negative bounding boxes to better reflect the 
proportion between false positive and false negative bounding boxes (as we found that an equal penalty caused the iterative training to become unstable very quickly).
%, likely due to the different proportion of false positive and false negative bounding boxes. 

For inference time, we found there may be overlapping tables that can be quite different 
in shape while having similar confidence levels. Thus, we choose a set of parameters ($\mu_5$, $\mu_6$, $\gamma_2$, $\epsilon$, $\delta$) to prioritize tables with the most tabular characteristics. In particular, 
we prioritize tables not having any cells within 2 lines of text outside the table ($\mu_5 = -20$ pixels) that are not contained already by other non-overlapping tables, 
while having many cells just inside the table, up to 0.25 of area (i.e., $\mu_6 = \{0.25*(x2-x1), 0.25*(y2-y1)\}$ pixels). 

\section{Cluster-based Algorithm for Generating Cell Structure}
\begin{algorithm}
\caption{Cell Boundary to Structure Cluster Algorithm}\label{euclid}
\begin{algorithmic}[1]
\Procedure{Preprocess Cell Bounding Boxes}{}
\For {$b$ in $cellboxes$} 
\If {not INTERSECT($b$, $textboxes$)} 
     \State DELETE $b$
\EndIf
\If {INTERSECT($b$, $textboxes$)} 
     \State $b$.bounding\_box = MAX($b$.bounding\_box, $textbox$.bounding\_box)
\EndIf
\If {INTERSECT($b$, $cellboxes$)} 
     \State $b$.bounding\_box = MAX($b$.bounding\_box, $cellbox$.bounding\_box)
\EndIf
\EndFor
\EndProcedure

\Procedure{Assign Cell Row and Column Location}{}
%\COMMENT{Expand cells horizontally to capture cases of unequal numbers of rows in each column}
\While {not INTERSECT($b$, $cellboxes$)}
    \State $b.x1 \gets b.x1 - 5$ 
    \State $b.x2 \gets b.x2 + 5$
\EndWhile
\For {$b$ in $cellboxes$}
    \State $num_{col} \gets$ MAX(CNT\_INTERSEC($b.midx$, $cellboxes$), $num_{col}$)
    \State $num_{row} \gets$ MAX(CNT\_INTERSEC($b.midy$, $cellboxes$), $num_{row}$)
\EndFor 
\State $alignment_x$, $alignment_y$ $\gets$ GET\_XY\_ALIGNMENT($cellboxes$)
\For {$b$ in $cellboxes$}
    \State $b.align_x$ $\gets$ ALIGN\_DATA($b.x1$, $b.midx$, $b.x2$, $alignment_x$)
    \State $b.align_y$ $\gets$ ALIGN\_DATA($b.y1$, $b.midy$, $b.y2$, $alignment_y$)
\EndFor
\State $col_{posx} \gets$ KMeans($cell_boxes.align_x$, $num_{col}$) 
\State $row_{pos}x \gets$ KMeans($cell_boxes.align_y$, $num_{row}$)
\For {$b$ in $cellboxes$}
    \State $b.col \gets$ ALIGN\_TO\_COL($b.align_x$, $col_{posx}$, $alignment_x$)
    \State $b.row \gets$ ALIGN\_TO\_ROW($b.align_y$, $col_{posy}$, $alignment_y$)
\EndFor
\EndProcedure

\Procedure{Assign Text Lines to Table}{}
\For {$b$ in $textboxes$} 
    \If {INTERSECT($b$, $cellboxes$)}
        \State $b.col \gets cellbox.col$
        \State $b.row \gets cellbox.row$
    \Else
        \State $b.col \gets$ ALIGN\_TO\_COL($b.align_x$, $col_{posx}$, $alignment_x$)
        \State $b.row \gets$ ALIGN\_TO\_ROW($b.align_y$, $col_{posy}$, $alignment_y$)
    \EndIf
\EndFor
\EndProcedure
% Need to add procedure for appending small case text
\Procedure{Split Cell Text Lines When Neighbor is Empty}{}
\For {$r$ in $num_{row}$}
     \For {$c$ in $num_{col}$}
         \If {IS\_EMPTY($r$, $c$)}
             \State $neighbor_{text} \gets$ GET\_CELLS($r-1$, $c$) + GET\_CELLS($r+1$, $c$)
             \For {$b$ in $neighbor_{text}$} 
                \State $b.col \gets$ ALIGN\_TO\_COL($b.align_x$, $col_{posx}$, $alignment_x$)
                \State $b.row \gets$ ALIGN\_TO\_ROW($b.align_y$, $col_{posy}$, $alignment_y$)
             \EndFor
         \EndIf
      \EndFor
\EndFor
\EndProcedure

\end{algorithmic}
\end{algorithm}

\section{Additional cell detection examples}
See Figures 2 and 3. 
\begin{figure*}[h]
\includegraphics[width=\textwidth]{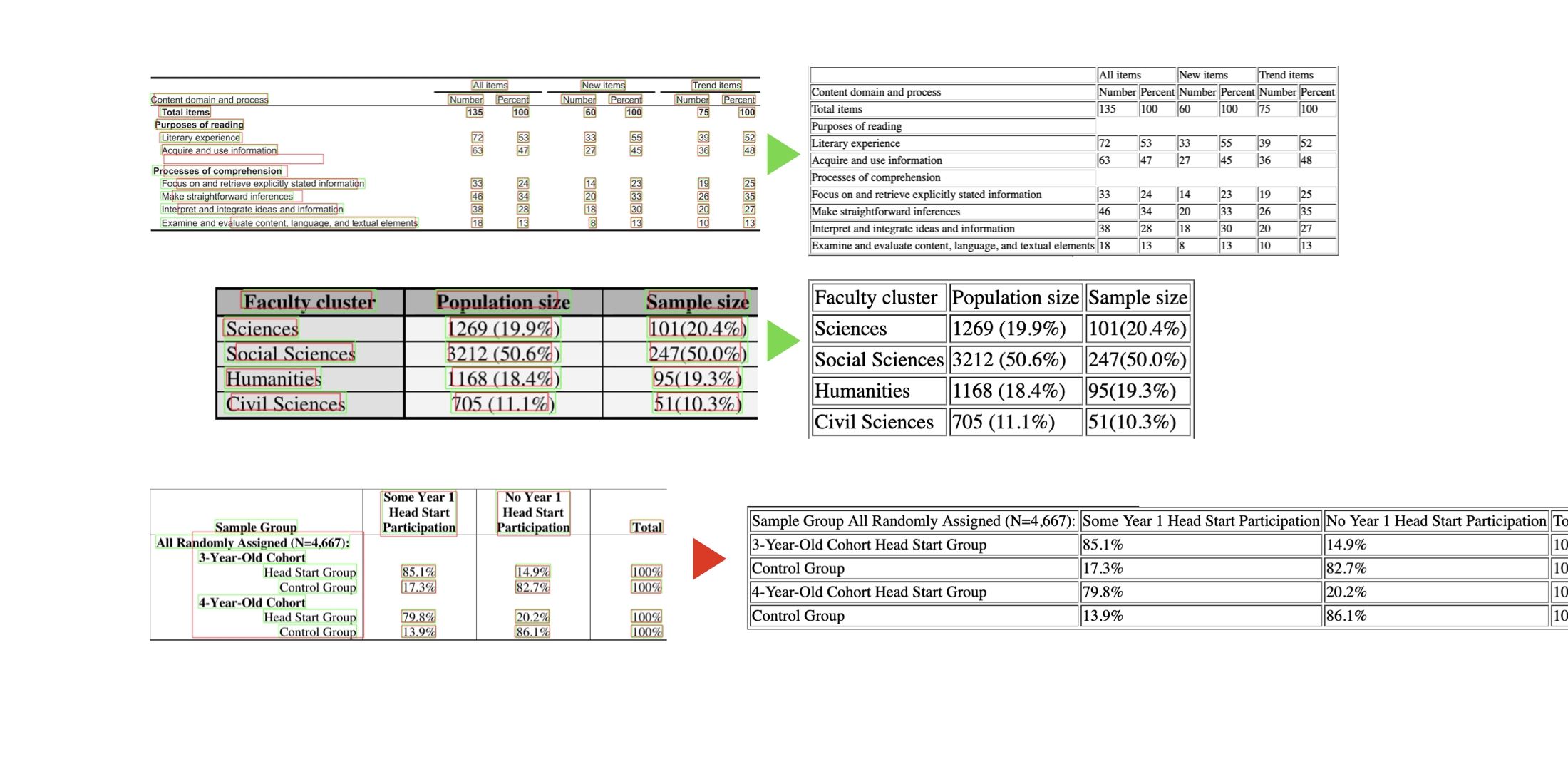}
\caption{Additional cell boundary to structure examples}
\end{figure*}

\begin{figure*}[h]
\caption{Example Cell detection errors}
        \begin{subfigure}[t]{0.4\textwidth}
        \includegraphics[width=\textwidth]{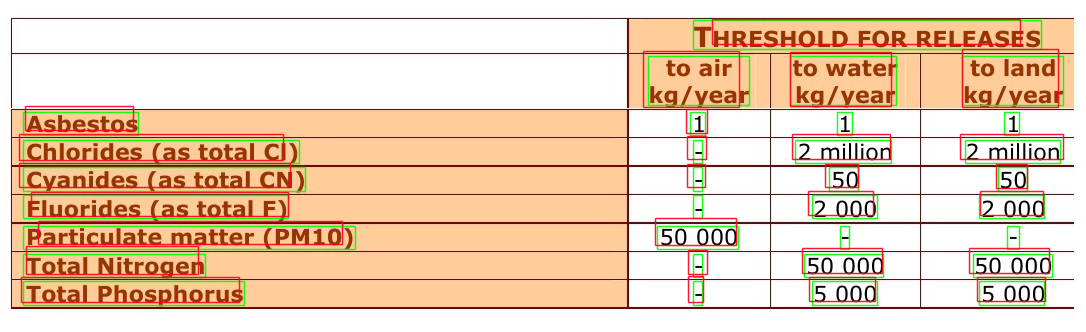}
        \caption{Correct cell detection}
    \end{subfigure}
    \begin{subfigure}[t]{0.4\textwidth} \includegraphics[width=\textwidth]{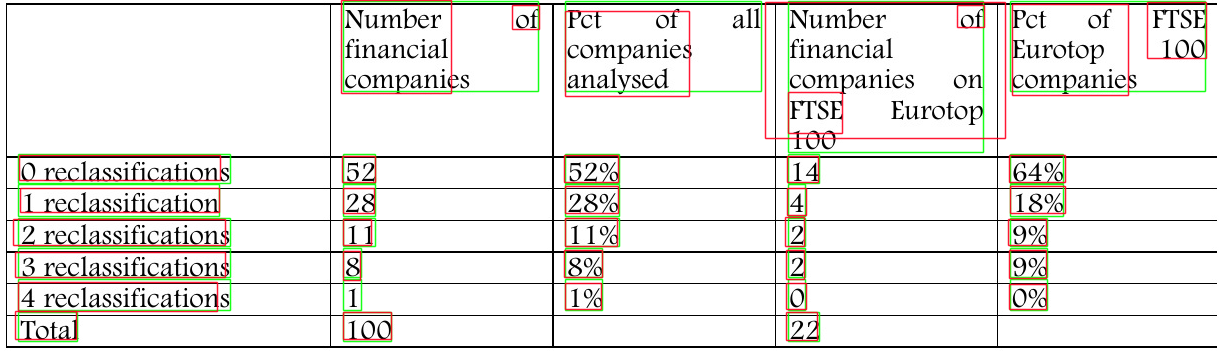}
        \caption{Oversplit cell detection}
    \end{subfigure}%
    ~ 
    ~
    \\
    \begin{subfigure}[t]{0.8\textwidth}
        \includegraphics[width=0.7\textwidth]{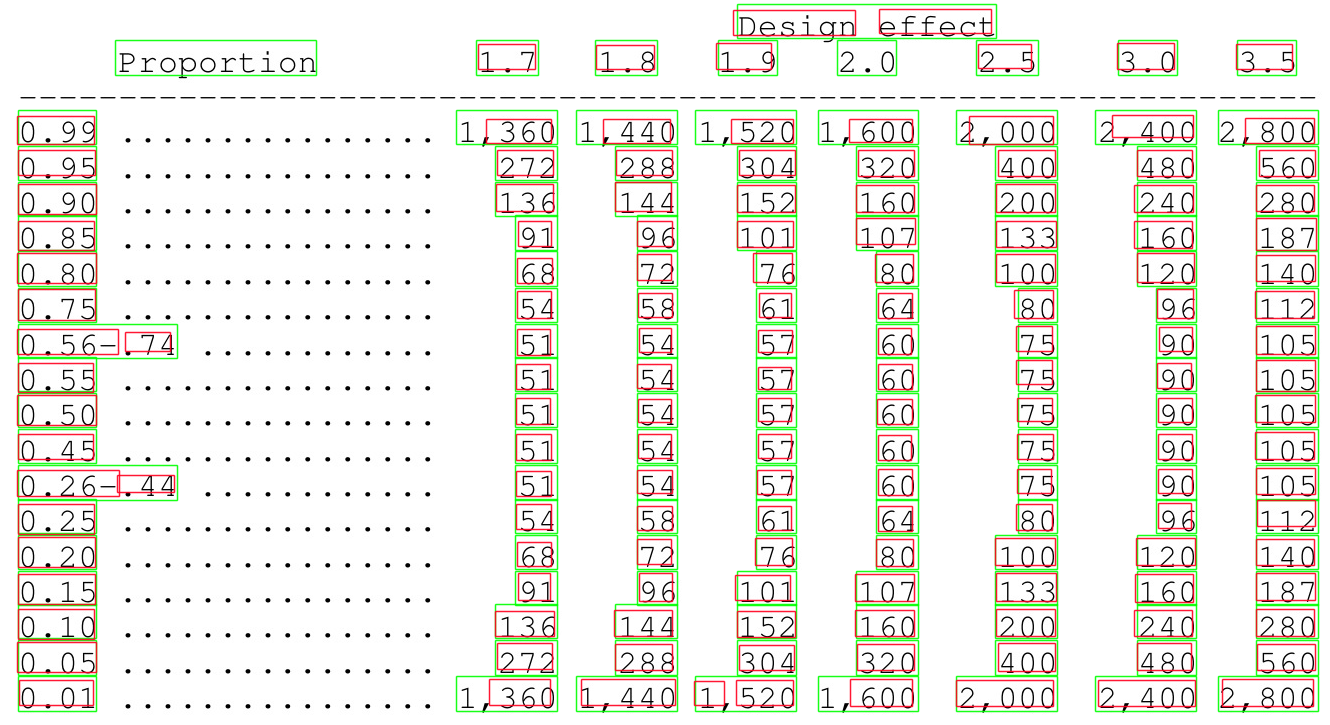}
        \caption{Missing Cell detection}
    \end{subfigure}
        \begin{subfigure}[t]{0.8\textwidth}
        \includegraphics[width=0.7\textwidth]{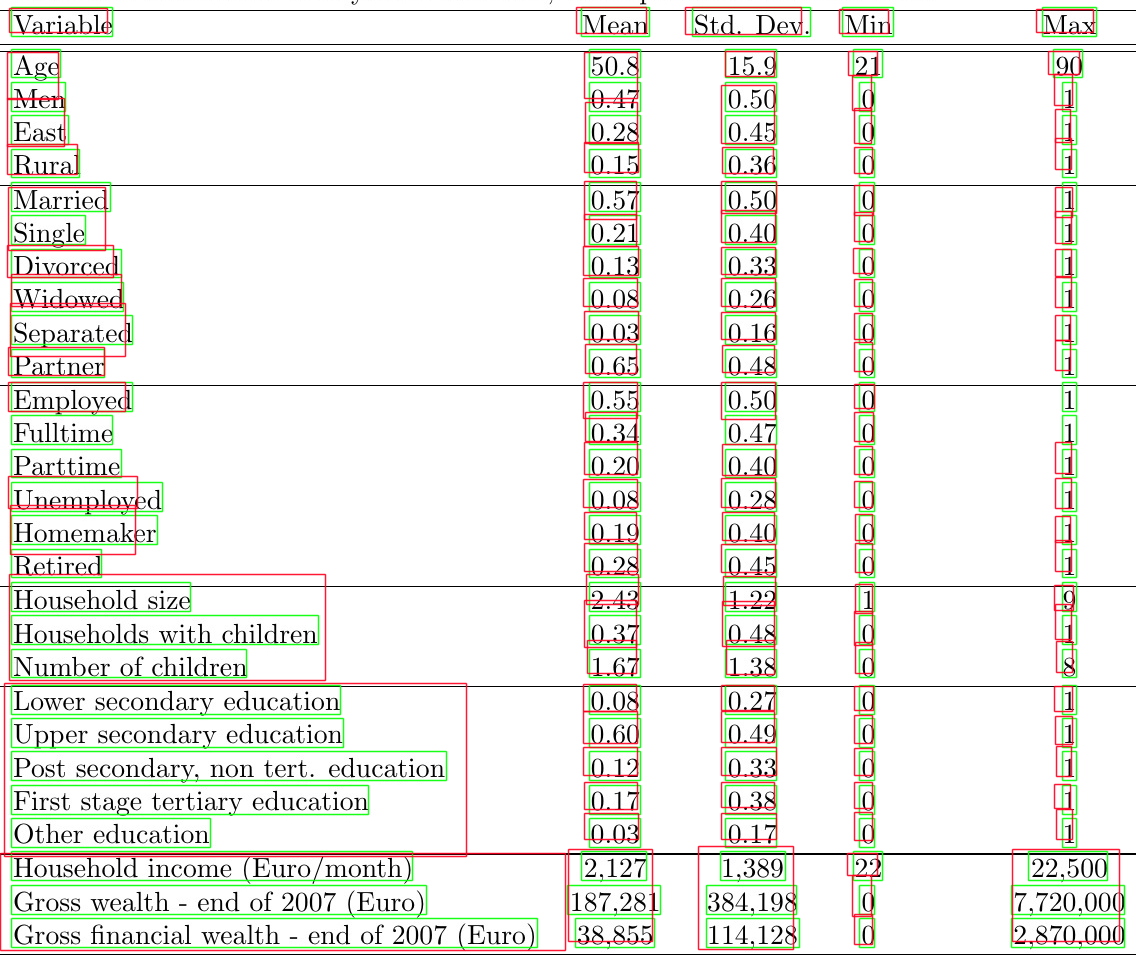}
        \caption{Overmerged cell detection}
    \end{subfigure}
\end{figure*}

\section{Detailed ICDAR13 Results}
See Tables 1 and 2. 
\begin{table}[H]
\caption{ICDAR 2013 table detection results with additional comparisons}
\centering
	\begin{tabular}{c|c|c||ccc|cc}
	    \hline
	    Category & Method & Input type & Recall & Precision & F1 & Cpt & Pu \\
	    \hline
	    {Commercial Softwares} & \textsl{FineReader} & PDF & 99.71 & 97.29 & 98.48 & 142 & 148\\
	    & \textsl{OmniPage} &  PDF & 96.44 & 95.69 & 96.06 & 141 & 130 \\
	    & \textsl{Nitro} &  PDF & 93.23 & 93,97 & 93.60 & 124 & 144 \\
	    & \textsl{Acrobat} &  PDF & 87.38 & 93.65 & 90.40 & 110 & 141 \\
	    \hline
	    {Non Deep Learning} 
	    & \textsl{ICST-Table}\cite{Fang2011ATD} &  PDF & 26.97 & 74.96 & 39.67  & 28 & 41\\
	    & \textsl{TableSeer}\cite{Liu2007TableSeerAT}  &  PDF & 33.35 & 88.36 & 48.64  & 0 & 29\\
	    & \textsl{Nurminen}\cite{Gbel2013ICDAR2T} &  PDF & 90.77 & 92.10 & 91.43  & 114 & 151 \\
        & \textsl{TABFIND}\cite{Silva2010Dis}  &  PDF & 98.31 & 92.92 & 95.54  & 149 & 137\\
        & \textsl{pdf2table}\cite{Yildiz2005pdf2tableAM}   &  PDF & 85.30 & 63.99 & 73.13 & 100 & 94 \\
        & \textsl{TEXUS}\cite{Rastan2015TEXUSAT}   &  PDF & 90.23 & 88.32 & 89.26  & 114 & 138 \\
	    \hline
	    {Deep Learning} 
	    & \textsl{Hao}\cite{Hao2016ATD} & Image & 97.24 & 92.15 & 94.63  & / &/ \\
	    & \textsl{DeepDeSRT}\cite{Schreiber2017DeepDeSRTDL} & Image & 96.15 & 97.40 & 96.77  & / & /\\
	    & \textsl{TableBank}\cite{Li2019TableBankTB}   & Image & / & / & 96.25  & / & /\\\hline
        Ours & GTE & Image & 99.77 & 98.97 & \bf{99.31} & 146 & 146\\
	    \hline
	\end{tabular}
	  	
  	\label{tab:td_results}
\end{table}
\begin{table}[H]
  \caption{Cell Structure results on ICDAR2013 with additional comparisons}
    \centering
    \begin{tabular}{c|c|c||ccc}
	    \hline
	    Category & Method  & GT Border? & Rec. & Prec. & F1  \\
	    \hline
	    {\centering Commercial Softwares} & \textsl{FineReader}  & N & 88.35 & 87.10 & 87.72 \\
	    & \textsl{OmniPage}  & N & 83.80 & 84.60 & 84.20 \\
	    & \textsl{Nitro} & N & 67.93 & 84.59 & 75.35 \\
	    & \textsl{Acrobat}  & N & 72.62 & 81.59 & 76.85 \\
	    \hline
	   {\centering Academic Systems} 
	      & \textsl{Nurminen}\cite{Gbel2013ICDAR2T} & N & 80.78 & 86.93 & 83.74  \\
	    & \textsl{TEXUS}\cite{Rastan2015TEXUSAT} & N & 84.23 & 81.02 & 82.59 \\
	            & \textsl{KYTHE}\cite{Gbel2013ICDAR2T} & N & 48.11 & 57.40 & 52.20 \\
        & \textsl{pdf2table}\cite{Yildiz2005pdf2tableAM} & N & 59.51 & 57.52 & 58.50 \\
         & \textsl{TABFIND}\cite{Silva2010Dis} & N & 70.52 & 68.74 & 69.62  \\\hline
        Ours	    & GTE & N & 92.72 & 94.41 & \bf{93.50}  \\\hline
	    {\centering Academic Systems}  & \textsl{Tensmeyer}\cite{tensmeyer2019deep} & Y &  94.64 & 95.89 & 95.26 \\
	    & \textsl{Nurminen}\cite{Gbel2013ICDAR2T} & Y &  94.09 & 95.12 & 94.60\\
	    & \textsl{Khan}\cite{khan2019table} & Y &  90.12 & 96.92 & 93.39\\

        & \textsl{TABFIND}\cite{Silva2010Dis} & Y &  64.01 &  61.44 & 62.70 \\

	    \hline
	   Ours & GTE & Y & 95.77 & 96.76 & \bf{96.24}\\
	    \hline
	    \hline
	\end{tabular}
    \label{tab:cstr_results}
\end{table}

\section{ICDAR19 evaluation metric ambiguities}
See Figure 4. 

\begin{figure}
    \centering
    \caption{The detected cell bounding boxes in the following images seem to be correct by eye and include all characters in the ground truth cell but has zero matches at IOU=0.9.}
    \includegraphics[width=\linewidth]{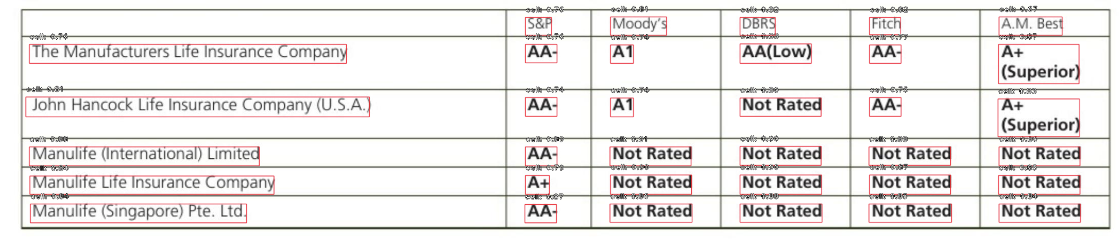}
    \includegraphics[width=\linewidth]{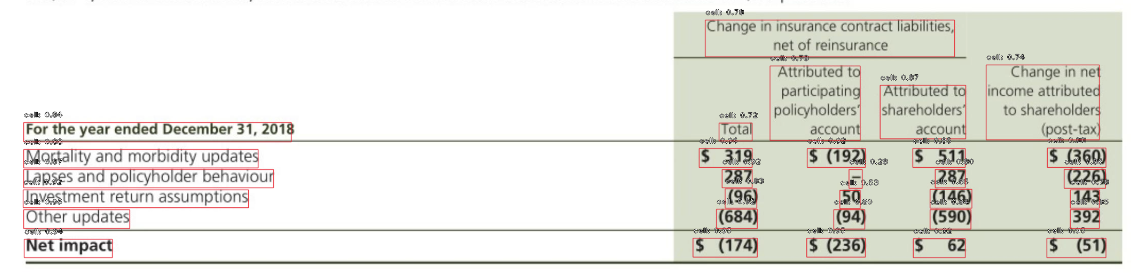}
    \includegraphics[width=\linewidth]{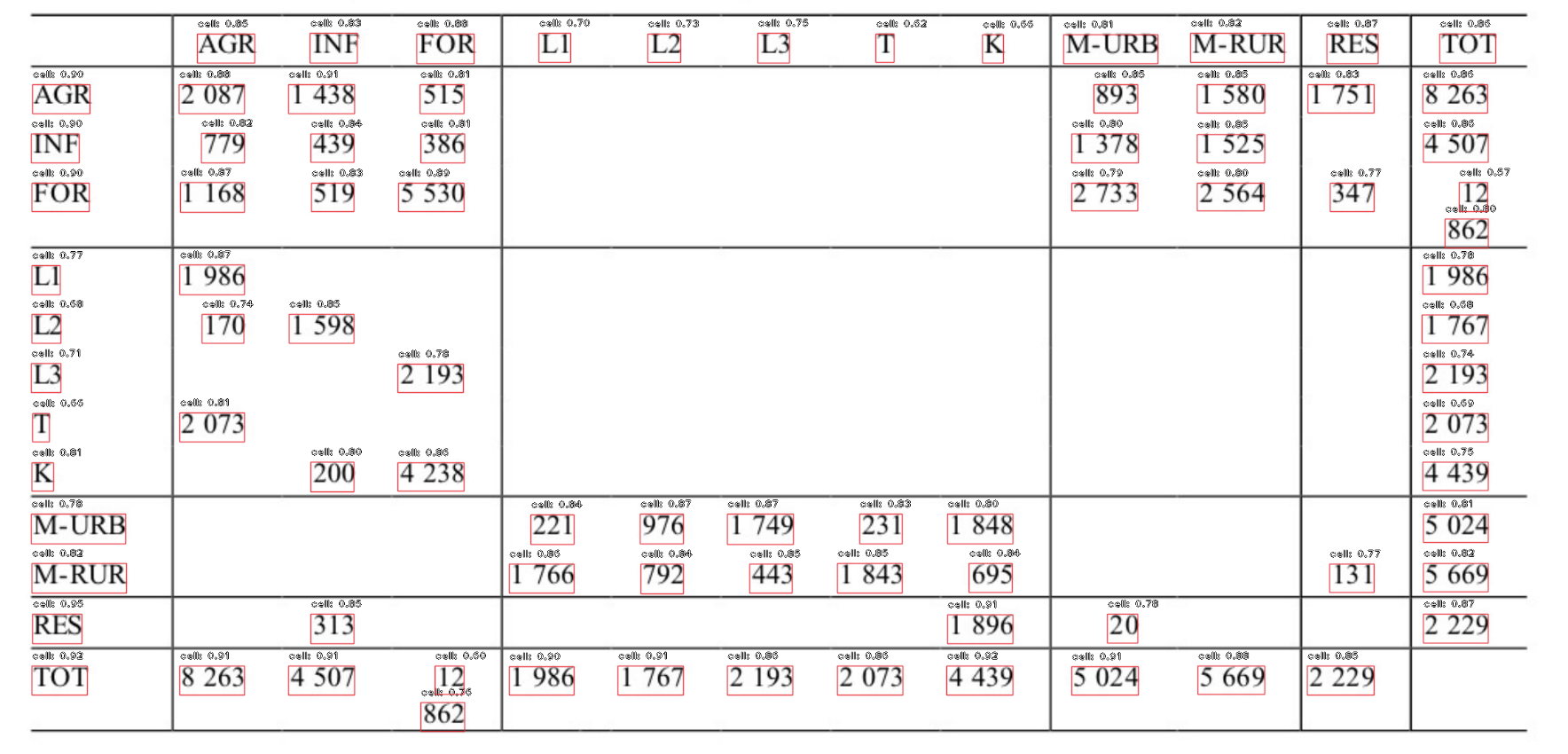}
    \label{fig:my_label}
\end{figure}

{\small
\bibliographystyle{ieee}
\bibliography{egbib}
}